%% file: main.tex
\pdfoutput=1

\documentclass[11pt]{article}

\usepackage[final]{acl}

\usepackage{times}
\usepackage{latexsym}
\usepackage{enumitem}

\usepackage{listings}
\usepackage{tabularx}
\usepackage{graphicx}
\usepackage{subcaption}
\usepackage{adjustbox}
\usepackage{amsmath}
\usepackage{algorithm}
\usepackage{algpseudocode}
\usepackage[longtable]{multirow}
\usepackage{longtable}
\usepackage{multicol}
\usepackage{caption}
\usepackage{subcaption}
\usepackage{float}
\usepackage{makecell}
\usepackage{amsmath}
\usepackage{amsthm}
\usepackage{amsfonts}
\usepackage{amssymb}
\graphicspath{{./image/}}
\usepackage{booktabs}
\usepackage{kotex} 
\usepackage{tabularray}
\usepackage{url}
\usepackage[normalem]{ulem}
\usepackage{tcolorbox}
\usepackage{pifont}
\usepackage{pdfpages}
\usepackage{color}
\usepackage{tikz}
\usepackage{circuitikz}
\usepackage{hyperref}
\usepackage{arydshln}
\usepackage{array}
\usepackage{needspace}

\lstdefinestyle{SQL}{
  language=SQL,
  basicstyle=\ttfamily\small,
  breaklines=true,
  keywordstyle=\color{blue},
  stringstyle=\color{red},
  commentstyle=\color{dkgreen},
  morecomment=[l]{\#},
  numbers=none 
}

\definecolor{Byzantine}{rgb}{0.74, 0.2, 0.64}
\definecolor{Mint}{rgb}{0.24, 0.71, 0.54}
\definecolor{Amaranth}{rgb}{0.9, 0.17, 0.31}
\definecolor{lightcyan}{RGB}{230,255,255}
\definecolor{lightyellow}{RGB}{255,255,224}
\definecolor{lightgreen}{RGB}{240,255,240}
\definecolor{ForestGreen}{RGB}{97,255,77}

\title{Towards Fully-Automated Materials Discovery via Large-Scale Synthesis Dataset and Expert-Level LLM-as-a-Judge}

\author{
 \textbf{Heegyu Kim\textsuperscript{1}$^*$},
 \textbf{Taeyang Jeon\textsuperscript{1}$^*$},
 \textbf{Seungtaek Choi},
 \textbf{Ji Hoon Hong\textsuperscript{3,4}},
 \textbf{Dong Won Jeon\textsuperscript{3,4}}, \\
 \textbf{Ga-Yeon Baek\textsuperscript{5}},
 \textbf{Gyeong-Won Kwak\textsuperscript{5}}, 
 \textbf{Dong-Hee Lee\textsuperscript{5}}, 
 \textbf{Jisu Bae\textsuperscript{5}},
 \textbf{Chihoon Lee\textsuperscript{5}},\\ 
 \textbf{Yunseo Kim\textsuperscript{5}},
 \textbf{Seon-Jin Choi\textsuperscript{5}},
 \textbf{Jin-Seong Park\textsuperscript{5}},
 \textbf{Sung Beom Cho\textsuperscript{3,4}},
 \textbf{Hyunsouk Cho\textsuperscript{1,2}},\\
 \textsuperscript{1}Department of Artificial Intelligence, 
 \textsuperscript{2}Department of Software and Computer Engineering,\\
 \textsuperscript{3} Department of Energy Systems Research,
 \textsuperscript{4} Department of Materials Science and Engineering,\\
 Ajou University, Suwon 16499, Republic of Korea\\
 \textsuperscript{5}Division of Materials Science and Engineering, Hanyang University, Seoul 04763, Republic of Korea \\
 \small{
   \textbf{Correspondence:} \href{mailto:hyunsouk@ajou.ac.kr}{hyunsouk@ajou.ac.kr}
 }
} 

\begin{document}
\maketitle

\def\thefootnote{*}\footnotetext{These authors contributed equally to this work}\def\thefootnote{\arabic{footnote}}

\input{00_macros}

\input{00_abs}
\input{01_intro}

\input{03_dataset}

\input{04_benchmark}

\input{05_llm_judge}

\input{06_exp}
\input{02_rel}

\input{07_con}

\input{11_limit}
\bibliography{custom}
\clearpage
\appendix

\input{appendix}

\end{document}

%% file: 00_macros.tex
\newcommand{\se}{{\it SE}}%
\newcommand{\eg}{{\it e.g.}}%
\newcommand{\ie}{{\it i.e.}}%
\newcommand{\etal}{{\it et al.}}%
\newcommand{\etc}{{\it etc}}%

\newcommand{\argmin}{\operatornamewithlimits{argmin}}
\newcommand{\argmax}{\operatornamewithlimits{argmax}}

\newcommand{\com}{\textcolor{red}}

\def\geotextual{{spatial-keyword}}
\def\geospatial{geo-spatial}
\def\PI{\mathcal{P}}
\newcommand{\XXP}[1]{{\PI(#1)}}
\def\XXQEO{\emph{$Q_1$}}
\def\kNN{\textsc{$k$NN}}
\def\XXD{\mathcal{D}}
\def\XXT{\mathcal{T}}
\newcommand{\XXDN}[0]{{D}}
\newcommand{\XXTN}[0]{{T}}
\def\Base{\textsc{Base}}
\def\TopK{\textsc{Top-$k$}}
\def\tag{{keyword}}
\def\Query{{Query}}
\newcommand{\ttag}[1]{{`#1'}}

\newcommand{\vanilla}{\textbf{Vanilla}}
\newcommand{\simple}{\textbf{FP-Simple}}

\newcommand{\mcal}[1]{{\cal{#1}}}
\newcommand{\calA}{\mbox{${\cal A}$}}
\newcommand{\calB}{\mbox{${\cal B}$}}
\newcommand{\calC}{\mbox{${\cal C}$}}
\newcommand{\calD}{\mbox{${\cal D}$}}
\newcommand{\calE}{\mbox{${\cal E}$}}
\newcommand{\calF}{\mbox{${\cal F}$}}
\newcommand{\calG}{\mbox{${\cal G}$}}
\newcommand{\calH}{\mbox{${\cal H}$}}
\newcommand{\calI}{\mbox{${\cal I}$}}
\newcommand{\calJ}{\mbox{${\cal J}$}}
\newcommand{\calK}{\mbox{${\cal K}$}}
\newcommand{\calL}{\mbox{${\cal L}$}}
\newcommand{\calM}{\mbox{${\cal M}$}}
\newcommand{\calN}{\mbox{${\cal N}$}}
\newcommand{\calO}{\mbox{${\cal O}$}}
\newcommand{\calP}{\mbox{${\cal P}$}}
\newcommand{\calQ}{\mbox{${\cal Q}$}}
\newcommand{\calR}{\mbox{${\cal R}$}}
\newcommand{\calS}{\mbox{${\cal S}$}}
\newcommand{\calT}{\mbox{${\cal T}$}}
\newcommand{\calU}{\mbox{${\cal U}$}}
\newcommand{\calV}{\mbox{${\cal V}$}}
\newcommand{\calW}{\mbox{${\cal W}$}}
\newcommand{\calX}{\mbox{${\cal X}$}}
\newcommand{\calY}{\mbox{${\cal Y}$}}
\newcommand{\calZ}{\mbox{${\cal Z}$}}

\newcommand{\LM}{\calL\calM}
\newcommand{\oursbench}{\textbf{AlchemyBench}}
\newcommand{\oursdatashort}{\textbf{OMG}}
\newcommand{\oursdatalong}{\textbf{Open Materials Guide}}
\newcommand{\testhi}{\textit{High Impact set}}
\newcommand{\testsi}{\textit{Standard Impact set}}

\newcommand{\db}{$D$}
\newcommand{\schema}{$S$}
\newcommand{\nlqset}{$X$}
\newcommand{\nlq}{$x$}
\newcommand{\sqlgt}{$Q_{gt}(x)$}
\newcommand{\sqlgen}{$Q_{gen}(x)$}
\newcommand{\exgt}{$R_{gt}(x)$}
\newcommand{\exgen}{$R_{gen}(x)$}

\newcommand{\todoc}[2]{{\textcolor{#1}{\textbf{#2}}}}
\newcommand{\todoorange}[1]{\todoc{orange}{\textbf{[[#1]]}}}
\newcommand{\hist}[1]{\todoorange{hist: #1}}

\newtcolorbox{defin}{enhanced,
	attach boxed title to top left={xshift=-4mm},boxrule=0pt,after skip=1cm,before skip=1cm,right skip=0cm,breakable,fonttitle=\bfseries,toprule=0pt,bottomrule=0pt,rightrule=0pt,leftrule=4pt,arc=0mm,skin=enhancedlast jigsaw,sharp corners,colframe=gree,colbacktitle=gre,boxed title style={
		frame code={ 
			\fill[gre](frame.south west)--(frame.north west)--(frame.north east)--([xshift=3mm]frame.east)--(frame.south east)--cycle;
			\draw[line width=1mm,gre]([xshift=2mm]frame.north east)--([xshift=5mm]frame.east)--([xshift=2mm]frame.south east);
			
			\draw[line width=1mm,gre]([xshift=5mm]frame.north east)--([xshift=8mm]frame.east)--([xshift=5mm]frame.south east);
			\fill[green!40](frame.south west)--+(4mm,-2mm)--+(4mm,2mm)--cycle;
		}
	}
}

%% file: 00_abs.tex
\begin{abstract}
Materials synthesis is vital for innovations such as energy storage, catalysis, electronics, and biomedical devices. Yet, the process relies heavily on empirical, trial-and-error methods guided by expert intuition. Our work aims to support the materials science community by providing a practical, data-driven resource. We have curated a comprehensive dataset of 17K expert-verified synthesis recipes from open-access literature, which forms the basis of our newly developed benchmark, AlchemyBench.
AlchemyBench offers an end-to-end framework that supports research in large language models applied to synthesis prediction. It encompasses key tasks, including raw materials and equipment prediction, synthesis procedure generation, and characterization outcome forecasting. We propose an LLM-as-a-Judge framework that leverages large language models for automated evaluation, demonstrating strong statistical agreement with expert assessments.
Overall, our contributions offer a supportive foundation for exploring the capabilities of LLMs in predicting and guiding materials synthesis, ultimately paving the way for more efficient experimental design and accelerated innovation in materials science.

\end{abstract}

%% file: 01_intro.tex
\section{Introduction}
\label{sec:intro}

Materials synthesis underpins advances in energy storage, catalysis, electronics, and biomedical devices~\cite{olivetti2020data}. Despite its importance, synthesis processes remain largely empirical, relying on trial-and-error approaches guided by expert intuition~\cite{merchant2023scaling}. This inefficiency highlights the need for systematic, data-driven approaches to predict synthesis workflows and optimize experimental design~\cite{huang2023application}.

\begin{figure}[!ht]
    \centering
    \includegraphics[width=\linewidth]{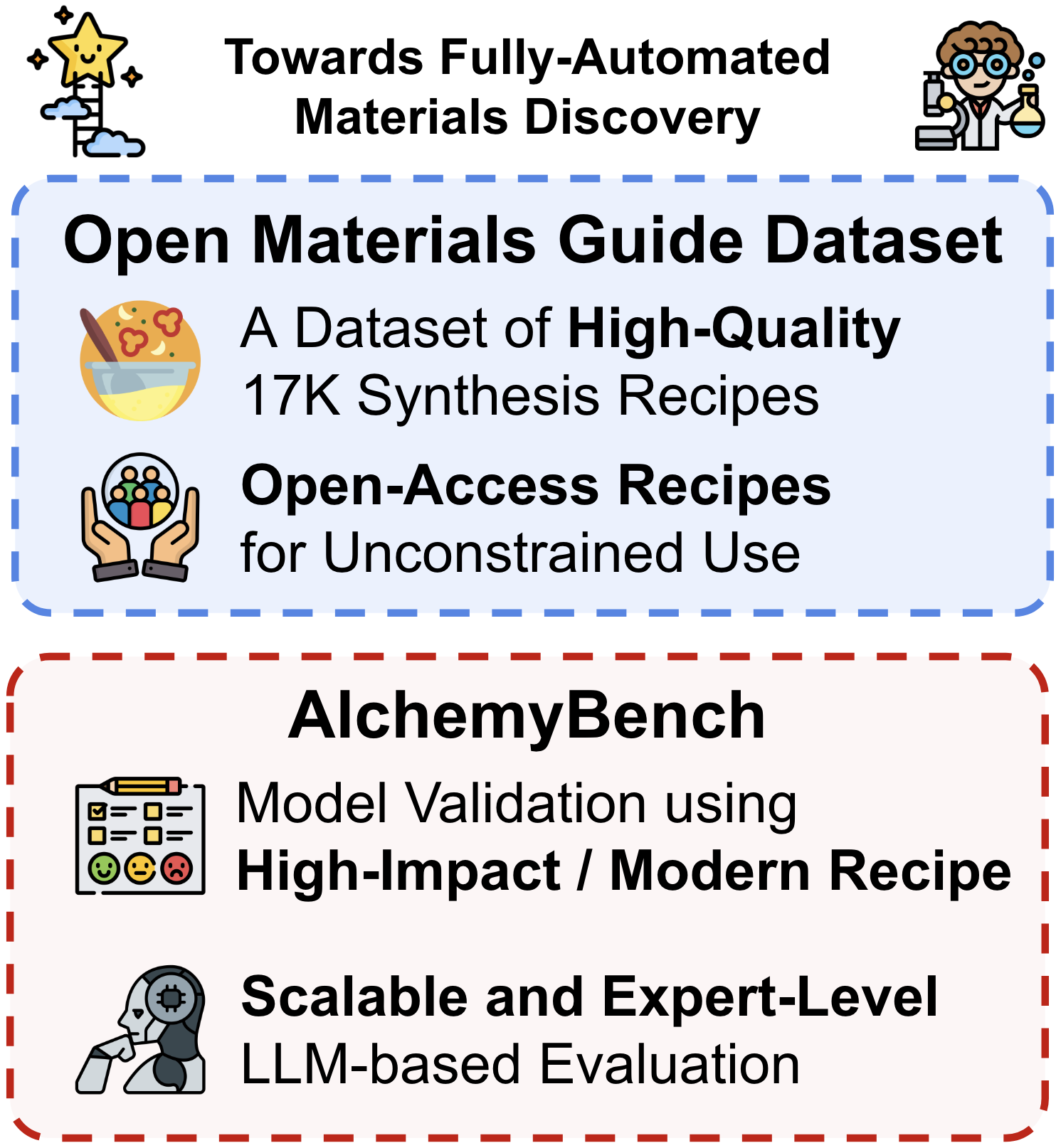}
    \caption{An overview of our contributions, featuring the Open Materials Guide Dataset for large-scale synthesis recipes and AlchemyBench for scalable, expert-level evaluation.}
    \label{fig:intro}
\end{figure}

\begin{figure*}[ht!]
    \begin{subfigure}{.48\textwidth}
    \centering
    \includegraphics[width=\linewidth]{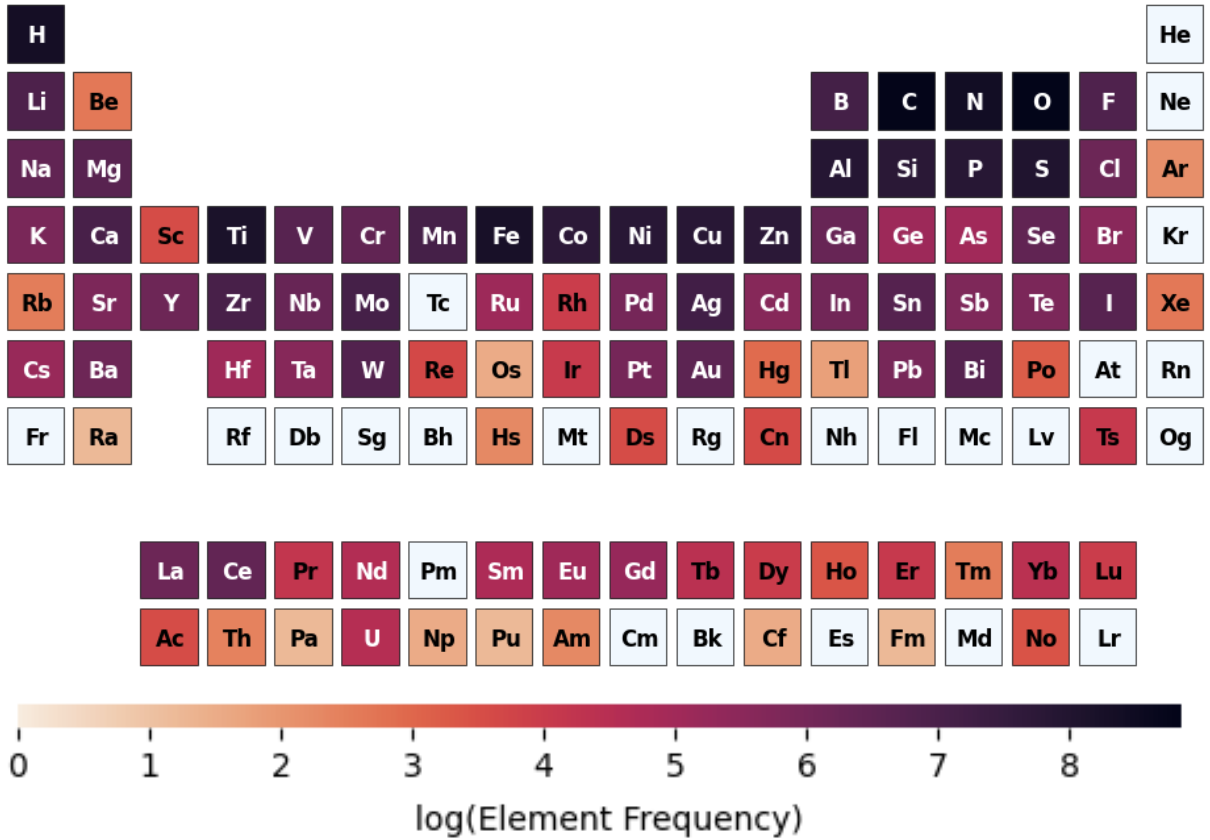}
    \caption{The periodic table of logarithmic frequency of elements.}
    \label{fig:data-element-distribution}
    \end{subfigure}
    \hfill
    \begin{subfigure}{.48\textwidth}
    \centering
    \includegraphics[width=\linewidth]{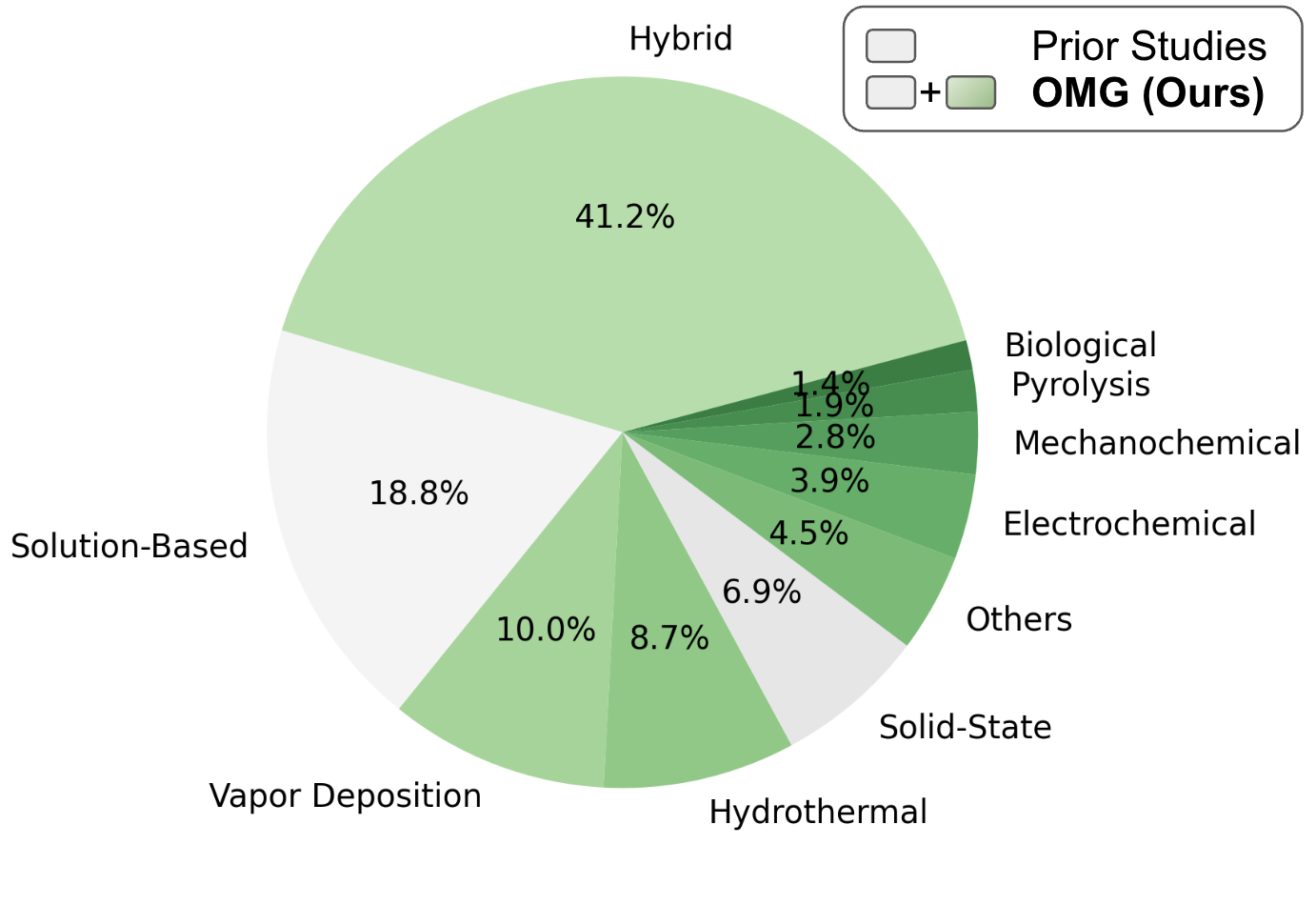}
    \caption{The distribution of synthesis techniques.}
    \label{fig:periodic-inorganic}
    \end{subfigure}

    \caption{The periodic table (left) demonstrates that \oursdatashort~covers diverse elements used in target materials, with darker colors indicating higher usage frequencies. A pie chart (right) illustrates the diversity of synthesis methods, highlighting the contributions of prior studies (white) and our dataset (white + green).}
    \label{fig:periodic-all}
\end{figure*}

Recent progress in machine learning and large language models has opened new avenues for extracting and generating synthesis procedures from unstructured scientific literature~\cite{song2023matsci,dunn2020benchmarking}. However, practical adoption is hampered by several challenges. Existing datasets are often small, domain-specific, and noisy, limiting model generalizability. Moreover, the absence of comprehensive benchmarks makes it difficult to assess the performance of synthesis prediction methods, while expert evaluations remain too costly and time-consuming for large-scale use.



To address these challenges, we introduce \oursdatalong~(\oursdatashort), a dataset comprising 17K high-quality, expert-verified synthesis recipes curated from open-access literature. This dataset is the foundation for our benchmark, \oursbench, which evaluates synthesis prediction across multiple facets—from inferring raw materials and recommending appropriate synthesis equipment to generating detailed procedural steps and forecasting suitable characterization techniques.

Additionally, we investigate an LLM-as-a-Judge framework to automate the evaluation process. Our systematic comparisons reveal a strong statistical agreement between LLM-based assessments and expert judgments, underscoring the potential of LLMs to serve as scalable, automated evaluators.

Our work makes the following key contributions:
\begin{itemize}
    \item \oursdatalong~(\oursdatashort), the most significant materials synthesis dataset, comprises 17K high-quality recipes from open-access literature. We demonstrated that various models can improve their performance with the proposed data-driven Retrieval-Augmented Generation (RAG)~\cite{lewis2020retrieval} experiments. From the improvement, we validate the applicability of our data.

    \item \oursbench, the first end-to-end benchmark for ML-driven synthesis prediction utilizing LLM-as-a-Judge, a scalable framework for evaluating synthesis predictions, demonstrating strong alignment with expert assessments. This framework enables automated benchmarking of synthesis prediction models, significantly reducing the reliance on costly and time-intensive expert evaluations while maintaining high evaluation reliability.
    
    \item \textbf{Extensive experimental insights} into model performance, identifying key challenges, potential capabilities, and future directions to utilize LLM for fully-automated materials synthesis.
    
\end{itemize}
To enhance reproducibility and accessibility, we release the dataset and code as an open-source resource for the research community\footnote{\url{https://github.com/HeegyuKim/AlchemyBench}}.

%% file: 03_dataset.tex
\section{Data Collection and Preparation}
\label{sec:data_collection }

\subsection{Motivation}
\label{subsec:motivation}

\begin{figure*}[t]
\centering
\includegraphics[width=\linewidth]{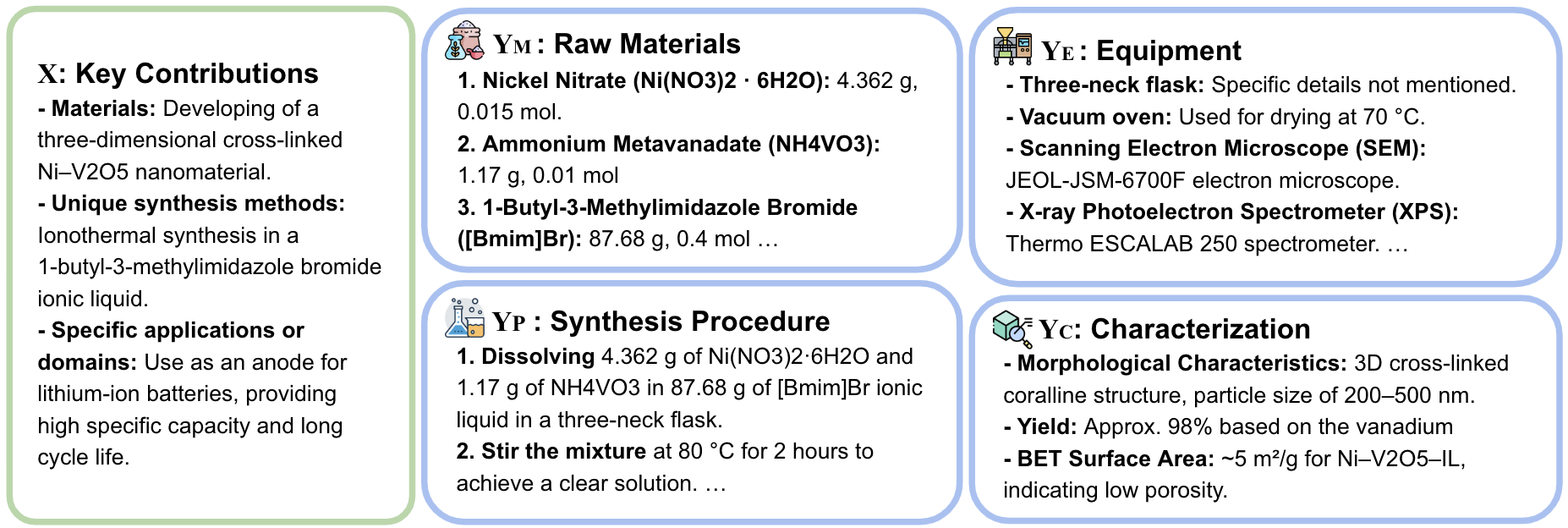}
\caption{An example of extracted recipe from~\citeauthor{zhao2020synthesis} demonstrates structured annotation of materials, equipment, procedures, and characterization methods.}
\label{fig:data-example}
\end{figure*}

Previous large-scale datasets for extracting synthesis procedures from materials science literature have faced several critical challenges~\cite{kononova2019text,wang2022dataset}. The most significant limitation involves common extraction errors—such as missing reagent concentrations, incorrect reaction temperatures, and misordered procedural steps—which have rendered many outputs unreliable for downstream synthesis prediction~\cite{sun2025critical}. We analyzed existing datasets and revealed that over 92\% of records in \citeauthor{kononova2019text} and 98\% in \citeauthor{wang2022dataset} lacked essential synthesis parameters (e.g., heating temperature, duration, mixing media). Additionally, these datasets are narrowly focused on a few synthesis techniques (such as solid-state and solution-based). At the same time, real-world materials innovation employs a broader range of specialized techniques~\cite{xu2023small}. Finally, copyright restrictions from commercial journals have limited the legal redistribution of textual synthesis procedures~\cite{authorsalliance2024tdm}.

To overcome these limitations, we propose~\oursdatashort~with three innovations: an LLM-driven parsing approach that improves extraction accuracy, a systematic collection covering more than ten distinct synthesis techniques (including vapor deposition, hydrothermal, and hybrid material systems), and the exclusive use of open-access publications to enable legal distribution of the dataset.

\label{subsec:quality_verification}

\subsection{Dataset Construction}
Our pipeline begins by retrieving 28,685 open-access articles from a pool of 400K search results using the Semantic Scholar API with 60 domain-specific search terms (e.g., ``solid state sintering process", ``metal organic CVD") recommended by domain experts. We convert PDFs to structured Markdown using \texttt{PyMuPDFLLM}~\cite{pymupdf4llm2024} and then employ GPT-4o in a multi-stage annotation process. First, articles are categorized based on whether they contain synthesis protocols, target materials, synthesis techniques, and applications. For articles confirmed to include synthesis procedures, the text is segmented into five key components, as illustrated in Figure~\ref{fig:data-example}:

\begin{itemize}
    \item \textbf{X}: A summary of the target material, synthesis method, and application.
    \item \(\mathbf{Y_M}\): Raw materials, including quantitative details.
    \item \(\mathbf{Y_E}\): Equipment specifications.
    \item \(\mathbf{Y_P}\): Step-by-step procedural instructions.
    \item \(\mathbf{Y_C}\): Characterization methods and results.
\end{itemize}

This systematic extraction yielded a dataset of 17,667 high-quality recipes (approximately a 62\% yield) covering 10 diverse synthesis methods. Figure~\ref{fig:periodic-all} demonstrates our dataset's broad coverage of materials systems and synthesis techniques. Detailed LLM prompts and search keywords are provided in Appendix~\ref{sec:appendix_dataset}.



\subsection{Quality Verification}
To ensure the accuracy of our automatically extracted recipes, we assembled a panel of eight domain experts from three institutions 
\footnote{Appendix~\ref{sec:appendix_annotation} describes the details about domain experts}. The experts manually reviewed a representative sample of ten recipes, evaluating them based on the following criteria:


\begin{itemize}
    \item \textbf{Completeness:} Capturing the full scope of the reported recipe (\textbf{X}, \(\mathbf{Y_M}\), \(\mathbf{Y_E}\), \(\mathbf{Y_P}\), and \(\mathbf{Y_C}\)).
    \item \textbf{Correctness:} Extracting critical details such as temperature values and reagent amounts accurately.
    \item \textbf{Coherence:} Retaining a logical, consistent narrative without contradictions or abrupt transitions.
\end{itemize}

Table~\ref{tab:human_verification} presents our expert evaluation results using a five-point Likert scale (1 = poor, 5 = excellent). To measure expert agreement, we computed the \textbf{Intraclass Correlation Coefficient (ICC)}~\cite{shrout1979intraclass}, utilizing a two-way mixed-effects model (ICC (3,k)) that quantifies agreement among evaluators, ensuring reliability in subjective scoring.
The extracted data exhibited high mean scores, but inter-rater reliability varied across criteria, particularly for articles with well-structured experimental sections.

\input{table/data/human_verification}
Completeness showed moderate agreement (ICC = 0.695), while correctness (ICC = 0.258) and coherence (ICC = 0.429) had lower agreement due to variations in naming conventions and missing characterization details.
Although the completeness score (4.2/5.0) was slightly lower than those for correctness (4.7/5.0) and coherence (4.8/5.0), correctness and coherence exhibited lower inter-rater reliability (ICC = 0.258 and 0.429, respectively), suggesting inconsistencies in how evaluators interpreted minor details. 
Variability in scores for correctness and coherence arose from differences in how evaluators weighted minor inconsistencies, such as variations in equipment naming or missing characterization information. Some considered these negligible, while others applied stricter criteria, underscoring the need for refined annotation guidelines.

While manual verification confirms the effectiveness of our extraction process, it cannot fully ensure consistent performance across the diverse range of synthesis procedures. In the following section (Section~\ref{sec:benchmark}), we present a structured evaluation framework for tasks such as raw materials and equipment inference, procedure generation, and characterization outcome forecasting.

%% file: table/data/human_verification.tex
\begin{table}
\centering
\caption{Data verification by eight domain experts.}
\label{tab:human_verification}
\begin{tabular}{l|cc} 
\toprule
\textbf{Criteria} & \textbf{Mean$_{~\sigma}$} & \textbf{ICC (3,k)}$_{p-\text{value}}$  \\ 
\hline
Completeness      & 4.2$_{~0.81}$             & 0.695$_{~0.00}$               \\
Correctness       & 4.7$_{~0.58}$             & 0.258$_{~0.23}$               \\
Coherence         & 4.8$_{~0.46}$             & 0.429$_{~0.10}$               \\
\bottomrule
\end{tabular}
\end{table}

%% file: 04_benchmark.tex
\section{AlchemyBench}
\label{sec:benchmark}

We present \oursbench, a comprehensive benchmark for evaluating materials synthesis prediction models. This framework addresses key challenges in synthesis recipe evaluation through structured tasks, expert-aligned metrics, and scalable assessment strategies.

\subsection{Motivation}
\label{subsec:motivation}

Evaluating synthesis predictions presents several fundamental challenges:

\begin{itemize}
    \item \textbf{Lack of Benchmarks:} No standardized evaluation framework exists, making it challenging to compare synthesis models systematically. Prior datasets lack critical synthesis parameters and structured ground truth labels, making meaningful comparisons difficult.
    \item \textbf{Limitations of Traditional Metrics:} Traditional metrics, such as BLEU~\cite{papineni2002bleu} and ROUGE~\cite{lin2004rouge} prioritize lexical overlap but fail to capture the procedural correctness of synthesis recipes. \citeauthor{na2023artificial} et al. introduced the Jaccard score to measure set overlap in synthesis procedures, yet it lacks sensitivity to sequential dependencies critical in procedural texts. BERTScore~\cite{zhang2019bertscore} improves contextual similarity measurement but struggles with domain-specific dependencies unique to materials synthesis. Moreover, these metrics do not account for experimental feasibility, limiting their applicability in real-world synthesis.
    \item \textbf{High Cost of Human Evaluation:} Expert-based assessments require significant time and resources, averaging 23 minutes per prediction ($\sigma=7.57$) in our experiment. This cost makes large-scale benchmarking impractical, requiring an automated evaluation system.
    \item \textbf{Scalability Requirements:} Large-scale benchmarking necessitates an automated yet reliable evaluation system, which LLMs can provide~\cite{gu2025surveyllmasajudge}. However, prior attempts to use LLMs for evaluation lacked systematic validation against human expert assessments in materials science, raising concerns about reliability.
\end{itemize}

\subsection{Task Definition}
\label{subsec:task_definition}

\oursbench~simulates real-world synthesis workflows, where models must predict the following components given input $\mathbf{X}$ (target material, synthesis method, application domain):

\begin{itemize}
    \item $\mathbf{P_M}$: Raw materials (e.g., reagents, solvents) with quantities.
    \item $\mathbf{P_E}$: Required equipment (e.g., furnace, autoclave).
    \item $\mathbf{P_P}$: Synthesis procedures (e.g., reaction steps, temperatures).
    \item $\mathbf{P_C}$: Characterization methods and expected outcomes.
\end{itemize}

\input{table/benchmark/criteria}
Predictions $\mathbf{P_X} = \{\mathbf{P_M,P_E,P_P,P_C}\}$ are evaluated against ground truth $\mathbf{Y_X} = \{\mathbf{Y_M,Y_E,Y_P,Y_C}\}$ using the LLM-as-a-Judge framework. Unlike prior benchmarks that rely on lexical similarity, \oursbench\ assesses procedural correctness and experimental feasibility. The evaluation criteria are described in Table~\ref{tab:judgment_criteria}.

The scoring function is computed as:

\[
\text{Score}(P_X, Y_X) = \frac{\sum_{i=1}^{N_C} C_i}{N_C}
\]

where $C_i$ represents the score for criterion $i$, and $N_C$ is the total number of evaluation criteria. These criteria were developed in collaboration with domain experts to ensure alignment with real-world synthesis evaluation.

\subsection{Dataset Splits and Distribution}
\label{subsec:dataset_splits}

We divided \oursdatashort~to three splits to ensure robust evaluation:

\begin{itemize}
    \item \textbf{Training Set:} 16,026 articles published before 2024.
    \item \textbf{Test - Standard Impact:} 1,472 articles (2024 and beyond) from journals with Impact Factor (IF) $<$ 10.
    \item \textbf{Test - High Impact:} 169 articles (2024 and beyond) from journals with IF $\geq$ 10.
\end{itemize}

The \textbf{temporal split} ensures that models are evaluated on \textit{unseen future research}, mitigating data contamination. Additionally, stratification by \textbf{journal impact} allows assessment of a model’s ability to process high-impact findings, often introducing novel and complex synthesis techniques. This split design evaluates both \textit{generalizability} and the ability to meet the rigorous standards of top-tier journals\footnote{Table~\ref{tab:high_impact_journals} describes the detailed list of high-impact journals utilized for our test-set split.}.

%% file: table/benchmark/criteria.tex
\begin{table*}[!ht]
\centering
\caption{Seven evaluation criteria used to evaluate synthesis recipes, categorized into materials, equipment, procedure, characterization, and overall score. Each criterion is rated on a 1–5 scale to reflect the quality and practicality of the predicted recipes.}
\label{tab:judgment_criteria}
\adjustbox{max width=\textwidth}{
\begin{tblr}{
  cells = {c},
  cell{4}{1} = {r=3}{},
  cell{7}{1} = {r=2}{},
  vline{2-3} = {1-10}{},
  vline{3} = {5-6,8}{},
  hline{1,10} = {-}{0.08em},
  hline{2-4,7,9-10} = {-}{},
}
\textbf{Category}         & \textbf{Criteria} & \textbf{Description}                                                                         \\
\textbf{{\includegraphics[height=0.4cm]{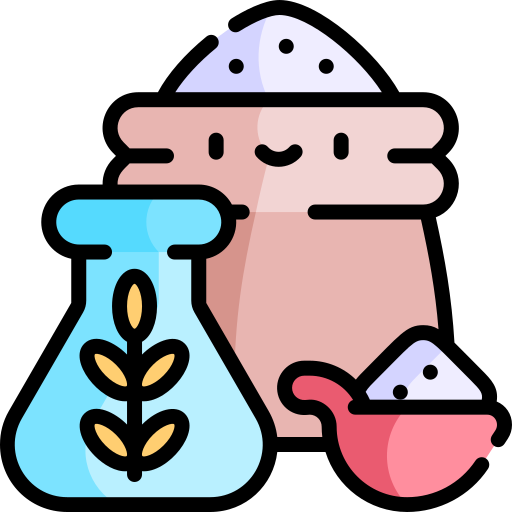}}~Materials}        & Appropriateness   & Are the selected materials suitable for the target synthesis?                                \\
\textbf{{\includegraphics[height=0.4cm]{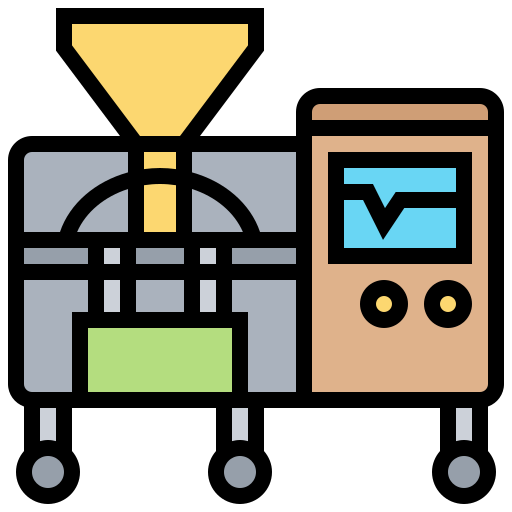}}~Equipment}        & Appropriateness   & Is the selected equipment suitable?                                                          \\
\textbf{{\includegraphics[height=0.4cm]{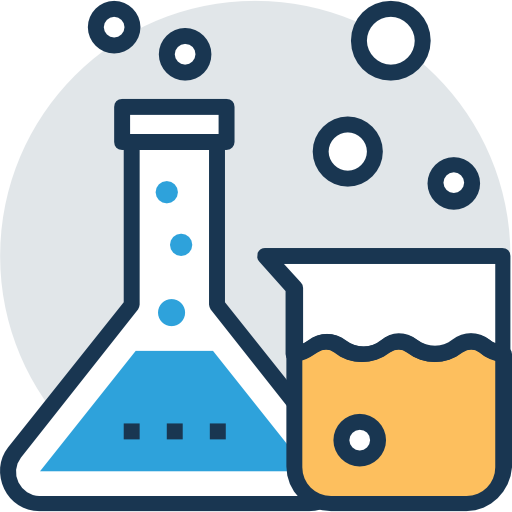}}~Procedure}        & Completeness     & Is the procedure well-organized and logically structured?                                    \\
                          & Similarity        & How closely does it match the ground truth procedure?                                        \\
                          & Feasibility       & Can this procedure be realistically executed in a lab?                                       \\
\textbf{{\includegraphics[height=0.4cm]{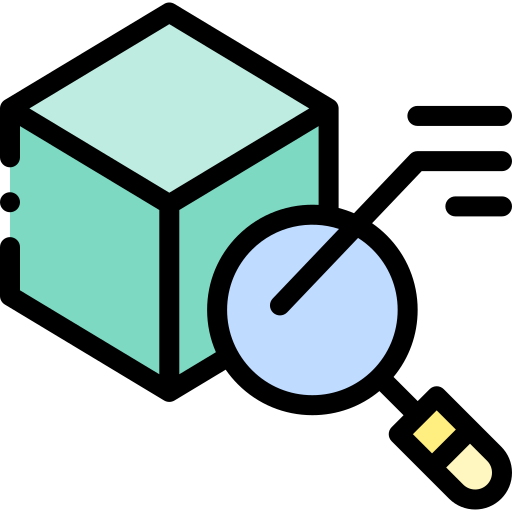}}~Characterization} & Appropriateness   & Are the methods and metrics suitable for validating the success of the synthesized material? \\
                          & Similarity        & How well do predicted properties match actual results?                                       \\
\textbf{Overall Score}    & -                 & Average score considering the recipe's overall quality and practicality.
\end{tblr}
}
\end{table*}

%% file: 05_llm_judge.tex
\section{LLM as a Judge}
\label{sec:reliability}

A reliable evaluation framework is essential for benchmarking synthesis prediction models. This section examines the alignment between LLM-based and human expert judgments, evaluating inter-rater agreement and assessing the effectiveness of LLMs as automated evaluators.

\subsection{Evaluation Metrics}
\label{subsec:evaluation_metrics}
We employ two metrics for evaluating the reliability of metrics: BLEU, ROUGE-L, BERTScore, and our LLM-as-a-Judge approach. \textbf{Pearson Correlation Coefficient} measures how closely LLM scores align with expert ratings on a continuous scale, capturing linear relationships. Finally, the \textbf{Spearman’s Rank Correlation} assesses rank-order consistency, beneficial when the relative ranking of recipes is more informative than absolute scores. 

\subsection{Human Expert Evaluation Setup}
\label{subsec:humaneval_setup}

Before evaluating whether the reliability of~\oursbench~assessment aligns with expert evaluations, we enlisted eight materials science researchers from three institutions to establish reliable ground truth. Each evaluator had prior experience in experimental synthesis and was selected based on their publication record and domain expertise. Experts independently assessed model-generated recipes using seven criteria (Table~\ref{tab:judgment_criteria}) on a 1–5 scale. To ensure high-quality assessments, we collected expert confidence scores and highlighted the agreement of one organization of three experts with the highest confidence levels on average (denoted as `High'). ICC(3,k) ensures the annotators' consensus's reliability. 

The dataset for evaluation included ten representative synthesis workflows selected by a senior materials scientist to ensure diversity. Prediction recipes were generated using two models (GPT-4o-mini and o1-mini), resulting in 20 unique predictions evaluated by human experts and LLM judges. Appendix~\ref{subsec:expert-protocol} and~\ref{sec:appendix_experiments} describe the experts annotation details and hyperparameters.

\subsection{Inter-Expert Agreement Analysis}
\label{subsec:inter_rater_analysis}
\input{table/experts/inter-raters-agreement}

The comparison between the High Confidence group and the All group in Table \ref{tab:inter-raters} highlights key differences in inter-rater reliability. The All group achieves higher ICC values for ``Material Appropriateness" (0.80) and ``Characterization Appropriateness" (0.78) compared to the High group (0.61 and 0.45, respectively), indicating better consensus among the broader panel for these criteria. However, the High group shows significantly stronger agreement on ``Procedure Feasibility" (ICC = 0.70) than the All group, which exhibits a negative ICC value (-0.58), suggesting inconsistencies in feasibility evaluations within the larger group. Both groups display similar reliability for ``Equipment Appropriateness" (ICC = 0.63). Overall, while larger panels may enhance agreement on straightforward criteria, smaller high-confidence subgroups provide more consistent evaluations for complex aspects like procedural feasibility.

\subsection{LLM-Expert Agreement Analysis}
\label{subsec:llm_expert_analysis}
\input{table/experts/llm-experts-agreement}

In Table~\ref{tab:llm-expert}, the traditional metrics (BLEU, ROUGE-L, and BERTScore) exhibit low or even negative correlations with the domain expert consensus regardless of the evaluator group, whereas the LLM-based scores consistently yield higher and statistically significant correlations. The values obtained for GPT-4o-mini, GPT-4o-Aug, and o3-mini (high) are notably higher in the high confidence subgroup—0.61, 0.80, and 0.62 respectively—compared to 0.45, 0.61, and 0.47 for the full panel, suggesting that evaluations from the more confident experts are more tightly aligned with these models. In contrast, GPT-4o-Nov shows a higher correlation with all eight experts (0.75) than with the high confidence subset (0.63), indicating that its performance remains robust even when considering a broader range of expert opinions. Overall, the comparison underscores the influence of expert group composition on evaluation outcomes and highlights the superior alignment of advanced LLM evaluators with expert assessments over traditional similarity metrics\footnote{Spearman correlation scores are described in Appendix~\ref{subsec:llm-expert-agreement-details}.}.

Our experiment confirms that LLM-generated scores correlate significantly better with expert assessments, supporting their use as scalable synthesis evaluators.






\subsection{Summary and Implications}
\label{subsec:summary_implications}

Our findings demonstrate that LLM-based evaluation provides a scalable and effective alternative to traditional synthesis assessment methods. GPT-4o-Aug exhibits strong agreement with expert ratings, outperforming traditional NLP metrics. 

However, challenges remain, as LLMs can be sensitive to ambiguous phrasing and domain-specific biases, affecting evaluation consistency. Future work should explore hybrid approaches integrating expert feedback with LLM scoring. Reinforcement learning from human feedback~\cite{ouyang2022training} (RLHF) and domain-specific fine-tuning~\cite{anisuzzaman2025fine} may improve alignment with expert reasoning.

Future work should investigate methods for mitigating biases and inconsistencies to enhance reliability, such as integrating expert validation into LLM-based evaluation pipelines. This study highlights the potential of LLMs as automated evaluators, paving the way for AI-driven, context-aware benchmarking frameworks in materials science.

%% file: table/experts/inter-raters-agreement.tex
\begin{table}[!h]
\centering
\caption{ICC (3,k) for each criterion. High denotes the highest confidence organization on average, and subscripts denote the $p$-value.}
\label{tab:inter-raters}
\adjustbox{max width=\linewidth}{
\begin{tblr}{
  column{2} = {r},
  column{3} = {r},
  cell{1}{2} = {c},
  cell{1}{3} = {c},
  vline{2} = {-}{},
  hline{1-2,10} = {-}{},
  hline{9} = {-}{dashed},
}
\textbf{Criteria}                & \textbf{High (3)} & \textbf{All (8)}  \\
Material Appropriateness         & 0.61$_{~0.01}$& 0.80$_{~0.00}$\\
Equipment Appropriateness        & 0.63$_{~0.00}$& 0.63$_{~0.00}$\\
Procedure Completeness           & 0.46$_{~0.05}$& 0.23$_{~0.19}$\\
Procedure Similarity             & 0.34$_{~0.14}$& 0.13$_{~0.31}$\\
Procedure Feasibility            & 0.70$_{~0.00}$& $-0.58_{~0.88}$ \\
Characterization Appropriateness & 0.45$_{~0.06}$& 0.78$_{~0.00}$\\
Characterization Similarity      & 0.37$_{~0.11}$& 0.45$_{~0.03}$\\
Overall Score (Average)          & 0.75$_{~0.00}$& 0.68$_{~0.00}$
\end{tblr}
}
\end{table}


%% file: table/experts/llm-experts-agreement.tex
\begin{table}[!h]
\centering
\caption{Pearson correlation coefficients between evaluation metrics and domain expert consensus for overall score. Subscripts denote the $p$-values. We set reasoning\_effort to high for o3-mini.}
\label{tab:llm-expert}
\adjustbox{max width=\linewidth}{
\begin{tblr}{
  column{2} = {r},
  column{3} = {r},
  cell{1}{2} = {c},
  cell{1}{3} = {c},
  vline{2} = {-}{},
  hline{1,9} = {-}{0.08em},
  hline{2} = {-}{},
  hline{5} = {-}{dashed},
}
\textbf{Model} & \textbf{High (3)}          & \textbf{All (8)}           \\
BLEU           & $-0.16_{~0.50}$          & $-0.23_{~0.34}$          \\
ROUGE-L        & $~0.06_{~0.80}$          & $-0.12_{~0.60}$          \\
BERTScore      & $-0.30_{~0.19}$          & $-0.24_{~0.31}$          \\
GPT-4o-mini    & $~0.61_{~0.00}$          & $~0.45_{~0.05}$          \\
GPT-4o-Aug     & \textbf{~0.80}$_{~0.00}$ & $~0.61_{~0.01}$          \\
GPT-4o-Nov     & $~0.63_{~0.00}$          & \textbf{~0.75}$_{~0.00}$ \\
o3-mini (high) & ~0.62$_{~0.02}$          & $~0.47_{~0.03}$          
\end{tblr}
}
\end{table}


%% file: 06_exp.tex
\section{Experiments}
\label{sec:experiment}


This section evaluates five LLMs on our benchmark using the LLM-as-a-Judge framework based on GPT-4o-Aug, analyzing performance across multiple metrics and the impact of retrieval-augmented generation (RAG).

\subsection{Experiment Setup}
\label{subsec:experiment_setup}

To comprehensively evaluate the models, we conducted experiments with the following setup:

\paragraph*{Base LLMs}
We evaluated four LLMs, including reasoning-based models (o3-mini) and general-purpose models (GPT-4o variants). The knowledge cutoff of these models is October 2023, minimizing potential data contamination in our test sets, which contain 2024 and beyond\footnote{The knowledge cutoff of OpenAI's models is described in \href{https://platform.openai.com/docs/models}{this documentation}.}. We prompt the LLM with a fixed one-shot example from our train set to predict all components ($\mathbf{P_X}$) well\footnote{Details about LLM prompt and hyperparameters are described in Appendix~\ref{sec:appendix_experiments}.}. Moreover, we varied the reasoning\_effort of o3-mini to ensure the effectiveness of reasoning effort in materials synthesis.

\paragraph*{Evaluation Framework}
Each model generated synthesis recipes for both the \testhi~and \testsi. Recipes were evaluated using our LLM-as-a-Judge method based on GPT-4o-Aug. The evaluation criteria focused on material appropriateness, procedural feasibility, and overall recipe quality.

\paragraph*{Retrieval-Augmented Generation (RAG)}
To evaluate the impact of retrieval on recipe generation, we implemented a RAG pipeline using OpenAI's text-embedding-3-large model~\cite{OpenAIEmbeedings2022}. For each input \(X\), we retrieved the top-\(K\) most similar recipes from the train set based on cosine similarity and included them as references in LLM prompts. We evaluated \(K = \{0, 1, 5, 10, 25\}\) to assess the effect of contextual information.

This experimental setup ensures a thorough evaluation of both baseline performance and improvements achieved through retrieval augmentation. Due to computational constraints, RAG experiments were conducted on three representative models (GPT-4o-Nov, GPT-4o-mini, and o3-mini) using only the \testhi.

\subsection{Insights from Experimental Results}
\label{subsec:insights}

\input{table/exp/base_results}

The experimental results provide valuable insights into the challenges and opportunities in materials synthesis prediction, structured around the following research questions:

\paragraph*{RQ1: Is High-Impact Set More Challenging than Standard-Impact Set?}
The results confirm that \testhi~is indeed more challenging than \testsi. Across models without GPT-4o-Nov, average scores on the \testhi~were generally lower than those on the \testsi. For example, o3-mini (high) achieved an average score of \(3.759 \pm 0.407\) on \testhi~compared to \(3.885 \pm 0.377\) on \testsi~(Table~5). This discrepancy highlights the increased complexity of High-Impact synthesis workflows, which often involve novel materials or cutting-edge techniques requiring greater reasoning and contextual understanding.

\paragraph*{RQ2: Does Increasing Reasoning Effort Improve Recipe Quality in Materials Science?}  
Materials science relies heavily on trial-and-error experimentation, making reasoning-based approaches particularly relevant for complex synthesis tasks. To explore this, we evaluated the o3-mini model using low, medium, and high reasoning efforts. As shown in Table~\ref{tab:base_results}, o3-mini (high) achieved the highest mean scores across both \textit{High-Impact} (\(3.759\)) and \textit{Standard-Impact} (\(3.885\)) sets, outperforming both its low and medium reasoning efforts and general-purpose models like GPT-4o-Nov. o3-mini (high) exhibited superior performane in step-by-step instructions compared to GPT-4o-Nov, which achieved a lower mean score (\(3.709\)) on the \testhi~despite matching o3-mini (high)'s maximum score (\(4.71\)). These findings demonstrate the importance of structured problem-solving capabilities in materials synthesis tasks.

\begin{figure}[ht!]
    \centering
    \includegraphics[width=\linewidth]{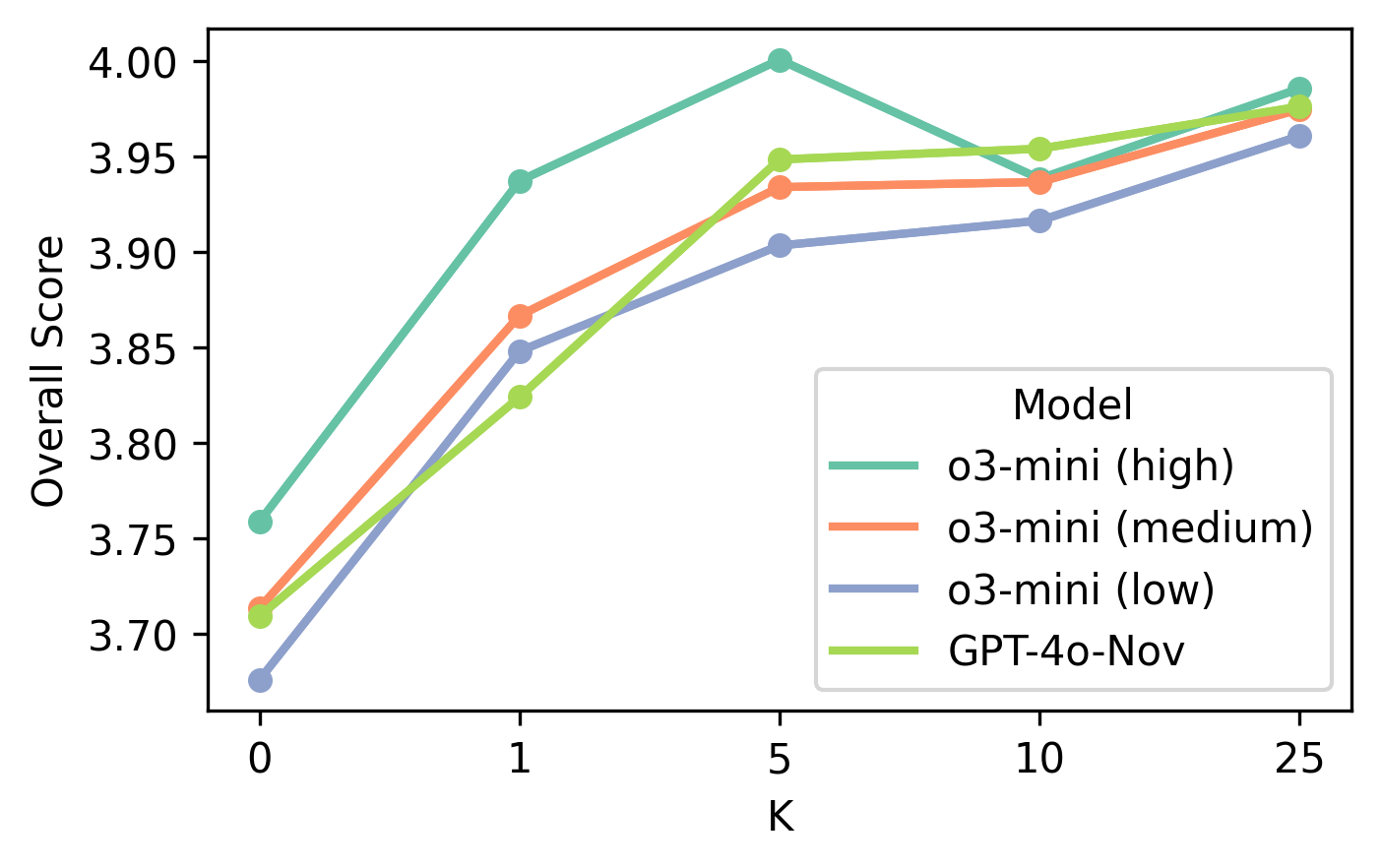}
    \caption{Impact of the Retrieval Augmented Generation (RAG) in \testhi.}
    \label{fig:rag_high_impact}
\end{figure}
\paragraph*{RQ3: Does Recipes of Similar Contributions Improve Prediction Performance?}
RAG with similar recipes significantly enhances model performance through domain-relevant examples. As shown in Figure~\ref{fig:rag_high_impact}, increasing \(K\) improves scores across all models, with different patterns between reasoning-based and general-purpose models.
For o3-mini (high), performance diminish after \(K=5\), reaching a maximum score of \(4.0\). In contrast, GPT-4o-Nov shows continuous improvement up to \(K=25\), achieving \(3.98\). The close performance between GPT-4o-Nov and o3-mini (high) suggests RAG can effectively bridge the gap between general-purpose and reasoning-based models.
These results highlight two key points: (1) RAG benefits general-purpose models that rely on external context, and (2) retrieval effectiveness depends on model architecture and training data quality. Our findings suggest that \(K=5\) provides the best trade-off between context enrichment and cognitive load, as additional references beyond this point yield minimal improvements.

%% file: table/exp/base_results.tex
\begin{table}[!ht]
\centering
\caption{Base LLMs' overall score evaluated by LLM-as-a-judge. Subscripts denote the standard deviation.}
\label{tab:base_results}
\adjustbox{max width=\linewidth}{
\begin{tabular}{lcccc}
\toprule
 & \multicolumn{2}{c}{\textbf{High Impact}} & \multicolumn{2}{c}{\textbf{Standard Impact}} \\
\textbf{Model} & {Mean$_{~\sigma}$} & Max & {Mean$_{~\sigma}$} & Max \\
\midrule
GPT-4o-mini & 3.238$_{~0.432}$ & 4.50 & 3.412$_{~0.412}$ & 4.71 \\
GPT-4o-Aug & 3.362$_{~0.405}$ & 4.50 & 3.508$_{~0.397}$ & 4.71 \\
GPT-4o-Nov & 3.709$_{~0.410}$ & \textbf{4.71} & 3.398$_{~0.397}$ & 4.71 \\
o3-mini (medium) & 3.714$_{~0.411}$ & 4.64 & 3.822$_{~0.387}$ & \textbf{4.80} \\
o3-mini (high) & \textbf{3.759}$_{~0.407}$ & \textbf{4.71} & \textbf{3.885}$_{~0.377}$ & \textbf{4.80} \\
\bottomrule
\end{tabular}
}
\end{table}


%% file: 02_rel.tex
\section{Related Work}

\paragraph*{Materials Synthesis Datasets}

Existing materials synthesis datasets, such as those focusing on solid-state~\cite{kononova2019text} and solution-based~\cite{wang2022dataset} methods, have provided valuable resources for machine learning applications. However, these datasets often suffer from issues of incompleteness and low quality, with many synthesis procedures lacking critical parameters necessary for reproducibility or predictive modeling. For instance, only 28\% of solid-state synthesis paragraphs yield complete reactions, and over 90\% of recipes are missing key parameters. These limitations hinder their utility in guiding novel synthesis workflows.

\paragraph*{LLM-based Generation for Materials Science}

Large language models (LLMs) have shown promise in accelerating materials discovery by automating hypothesis generation~\cite{kumbhar2025hypothesisgenerationmaterialsdiscovery}, property prediction~\cite{chiang2024llamplargelanguagemodel}, and evaluation~\cite{mishra2024llamat}. However, their effectiveness is constrained by the lack of high-quality domain-specific datasets and the need for retrieving or fine-tuning to handle complex synthesis workflows. Our work addresses these gaps and introduces a large-scale dataset and benchmark tailored to the real-world synthesis workflow, enabling rigorous evaluation of LLM capabilities in materials science.

%% file: 07_con.tex
\section{Conclusion}
\label{sec:conclusion}

This study presents a comprehensive benchmark for evaluating LLMs in materials synthesis prediction, addressing key challenges in data-driven materials science. By curating a large-scale dataset and designing tasks that mirror real-world synthesis workflows, we provide a robust framework to assess model capabilities in raw materials selection, equipment inference, procedure generation, and characterization prediction. Our experiments reveal the potential of reasoning-based models, such as o3-mini, outperforming general-purpose models like GPT-4o variants in generating coherent and feasible synthesis recipes. Furthermore, integrating retrieval-augmented generation (RAG) enhances recipe quality by grounding predictions in domain-relevant examples, with optimal performance gains observed at \(K=5\). These findings underscore the importance of combining advanced reasoning architectures with adaptive retrieval strategies for materials science tasks, laying the foundation for interdisciplinary innovation and accelerating progress in data-driven and fully-automated materials discovery.

%% file: 11_limit.tex
\section{Limitations}
\label{sec:limitations}

While our benchmark represents a significant step toward integrating LLMs into materials synthesis, several limitations remain. First, the dataset, derived from open-access articles, may exhibit biases in domain coverage, overrepresenting fields like battery materials while underrepresenting others like biomaterials. Second, GPT-4o for recipe extraction and evaluation introduces potential inaccuracies and biases, particularly in complex or ambiguous texts. Third, while practical, reliance on LLM-based scoring may oversimplify the nuanced requirements of tasks like procedure generation and characterization prediction. Additionally, the sequential dependencies between tasks (e.g., precursor prediction influencing procedure generation) pose challenges for current models, which may overfit dataset-specific patterns rather than learning generalizable principles. Finally, the lack of interpretability in model outputs limits their applicability in critical experimental workflows. Addressing these issues through improved data curation, expanded evaluation frameworks, and developing more interpretable models will be vital for future progress in this domain.
 

%% file: appendix.tex
\input{table/data/high_impact_journals}

\begin{figure}
    \centering
    \includegraphics[width=\linewidth]{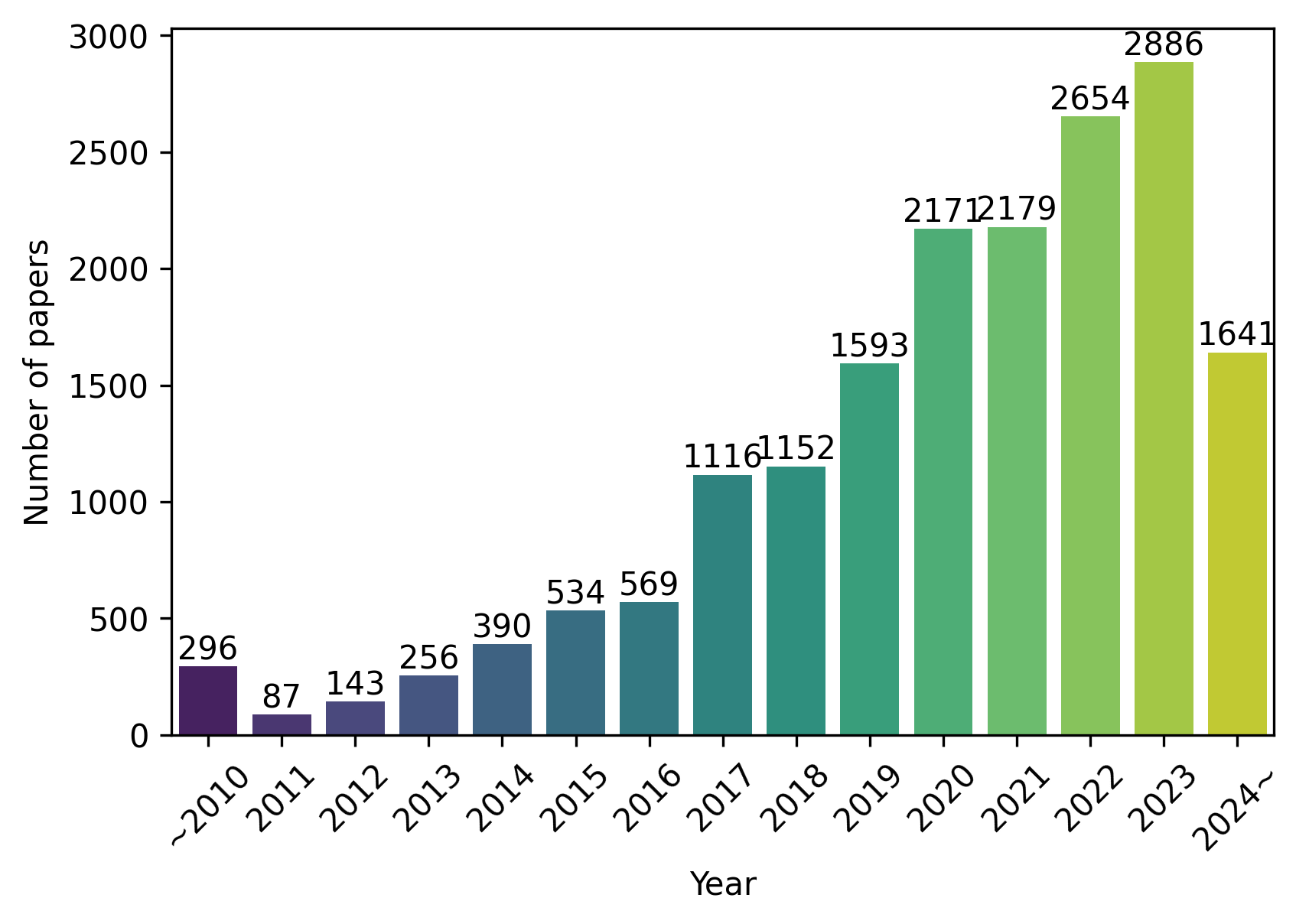}
    \caption{Yearly distribution of collected material synthesis papers}
    \label{fig:data-yearly-distribution}
\end{figure}

\appendix
\section{Additional Dataset Information}
\label{sec:appendix_dataset}

This section provides extended details on dataset statistics and collection methodology that were not included in the main text for brevity.

\subsection{Keyword Selection Rationale}

Figure~\ref{fig:prompt_search_keywords} describes the search keywords to retrieve 400K articles using Semantic Scholar API~\cite{semanticscholar2023}. We collect these keywords guided by our eight domain experts.

\subsection{Downloading PDFs}
We downloaded open-access papers exclusively from the following six publishers, most frequent in our retrieval result: \href{https://pubs.rsc.org}{pubs.rsc.org}, \href{https://www.mdpi.com}{mdpi.com}, \href{https://www.nature.com}{nature.com}, \href{https://link.springer.com}{link.springer.com}, \href{https://pubs.acs.org}{pubs.acs.org}, \href{https://onlinelibrary.wiley.com}{onlinelibrary.wiley.com}.

\subsection{Dataset Details}
\begin{figure}[ht!]
    \centering
    \begin{subfigure}{\linewidth}
        \centering
        \includegraphics[width=\linewidth]{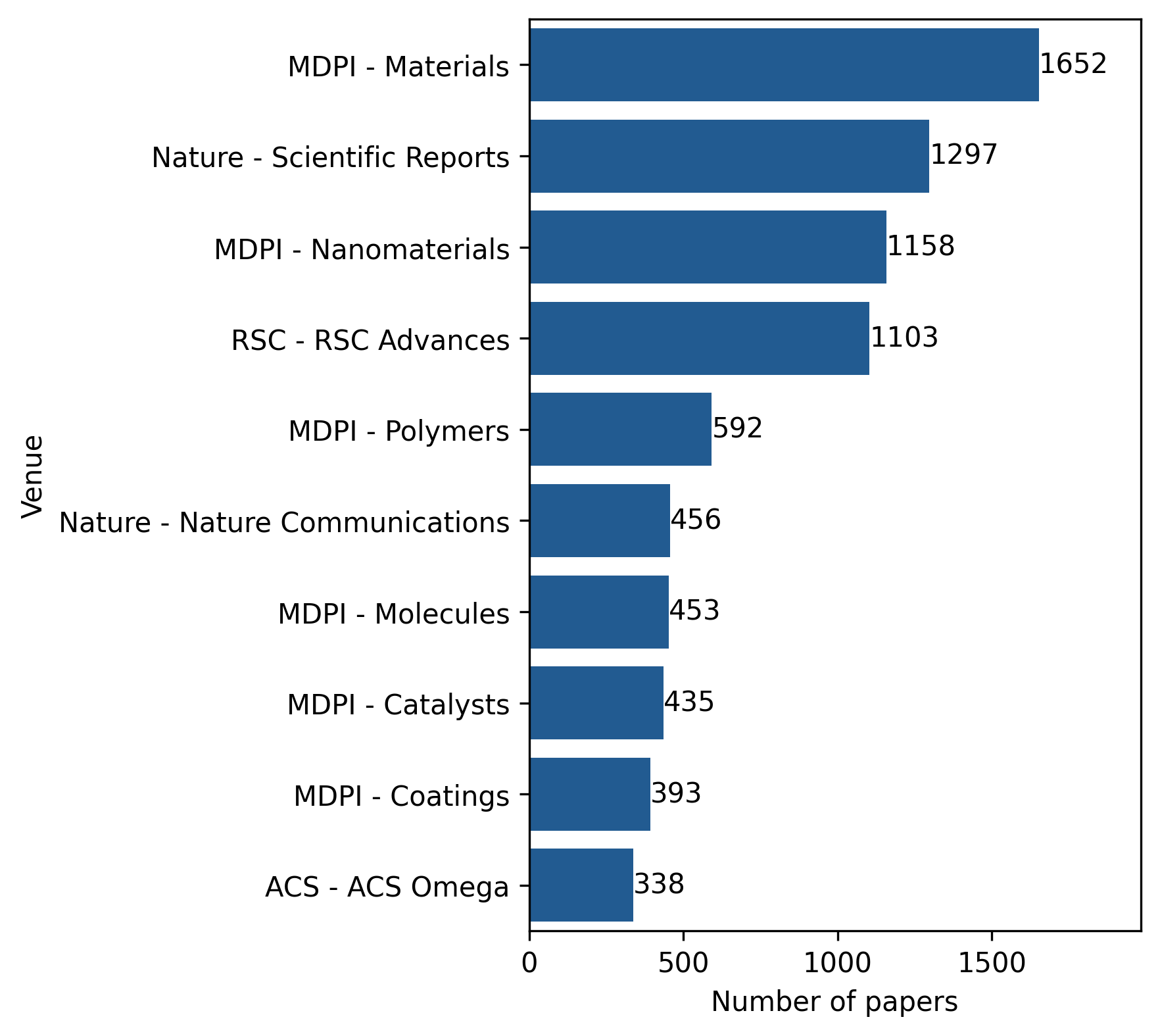}
        \caption{A venue distribution of the train set}
        \label{fig:train-venue-distribution}
    \end{subfigure}
    
    \begin{subfigure}{\linewidth}
        \centering
        \includegraphics[width=\linewidth]{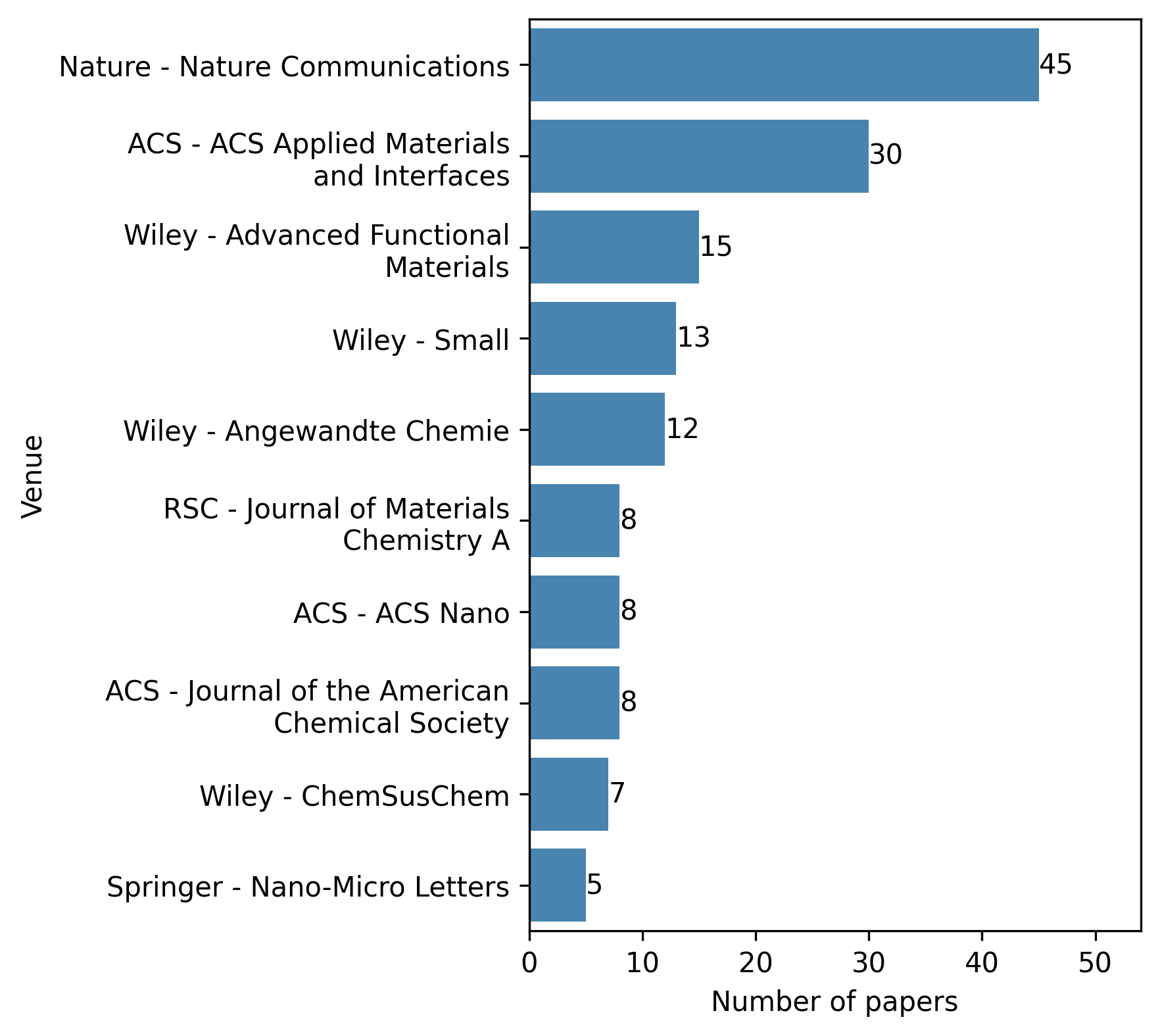}
        \caption{A venue distribution of the test high-impact set}
        \label{fig:test-high-impact-venue-distribution}
    \end{subfigure}
    
    \begin{subfigure}{\linewidth}
        \centering
        \includegraphics[width=\linewidth]{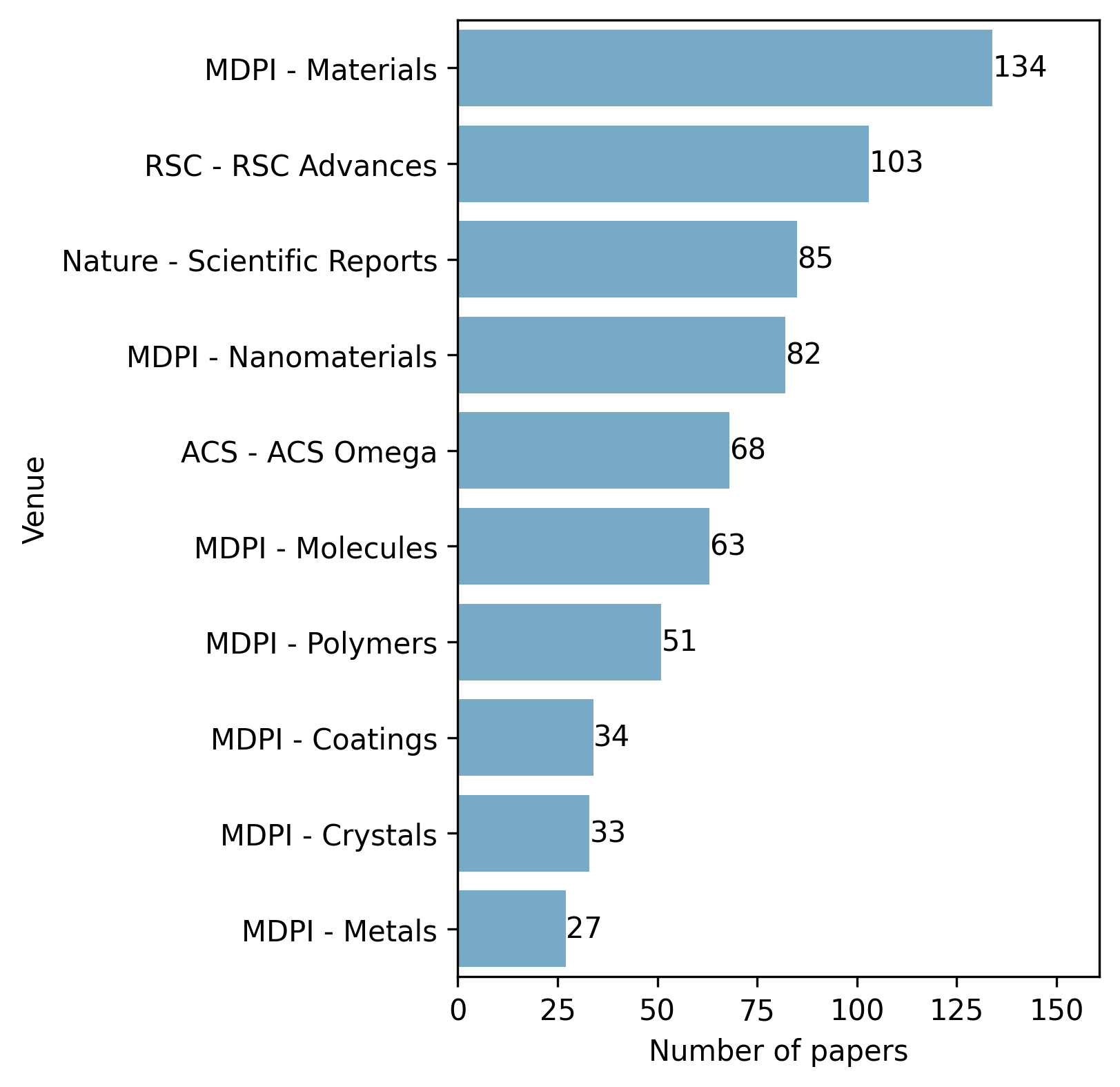}
        \caption{A venue distribution of the test standard impact set}
        \label{fig:test-standard-impact-venue-distribution}
    \end{subfigure}

    \caption{Venue distributions across datasets: (a) training set, (b) test high-impact set, and (c) test standard impact set. The distributions illustrate the diversity and focus of venues in each subset.}
    \label{fig:venue-distributions}
\end{figure}

Figure~\ref{fig:train-venue-distribution},~\ref{fig:test-standard-impact-venue-distribution}, and~\ref{fig:test-high-impact-venue-distribution} demonstrate the distributions of venue for train, test-standard-impact, and test-high-impact respectively. Table~\ref{tab:high_impact_journals} describes the high-impact venues (IF $\geq 10$) that at least ten papers are included in~\oursdatashort. Figure~\ref{fig:data-yearly-distribution} demonstrates the dataset distribution of the published year, indicating the latest data, 2020 and beyond, accounts for a large percentage.

\vspace{2em}
\section{Annotation Pipeline and Quality Checks}
\label{sec:appendix_annotation}

Here, we elaborate on the annotation workflow and the validation methods used to ensure the reliability of extracted recipes.

\subsection{GPT-4 Prompts for Classification and Extraction}
Figure~\ref{fig:prompt_paper_categorization} demonstrates the prompt to categorize the literature and Figure~\ref{fig:prompt_paper_extraction} demonstrates the prompt to extract the synthesis recipe from the literature.

\subsection{Expert Review Protocol}
\label{subsec:expert-protocol}

\begin{figure*}
    \centering
    \includegraphics[width=\textwidth]{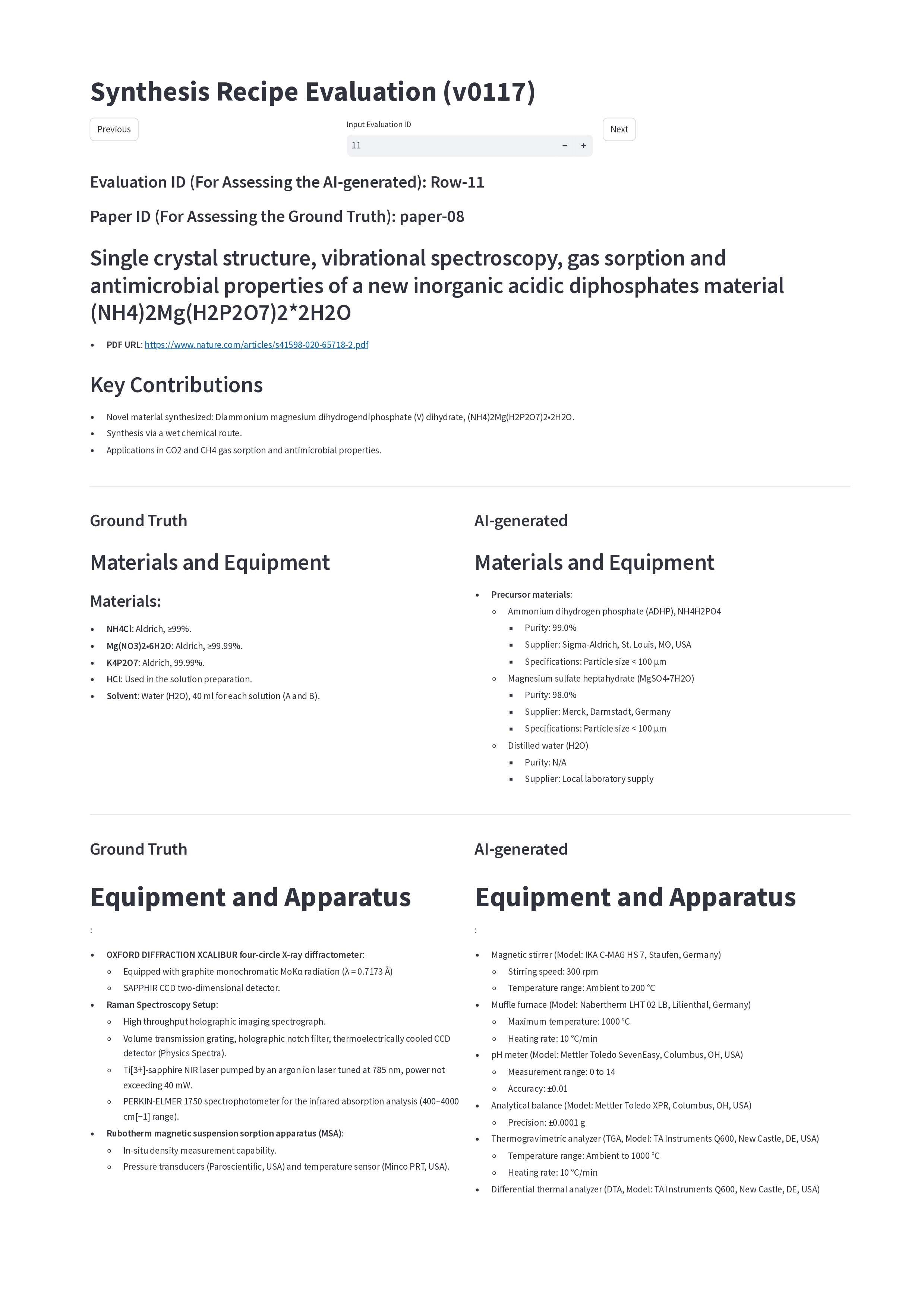}
    \caption{A web UI screenshot for domain experts annotation (1/2).}
    \label{fig:experts-webui-1}
\end{figure*}

\begin{figure*}
    \centering
    \includegraphics[width=\textwidth]{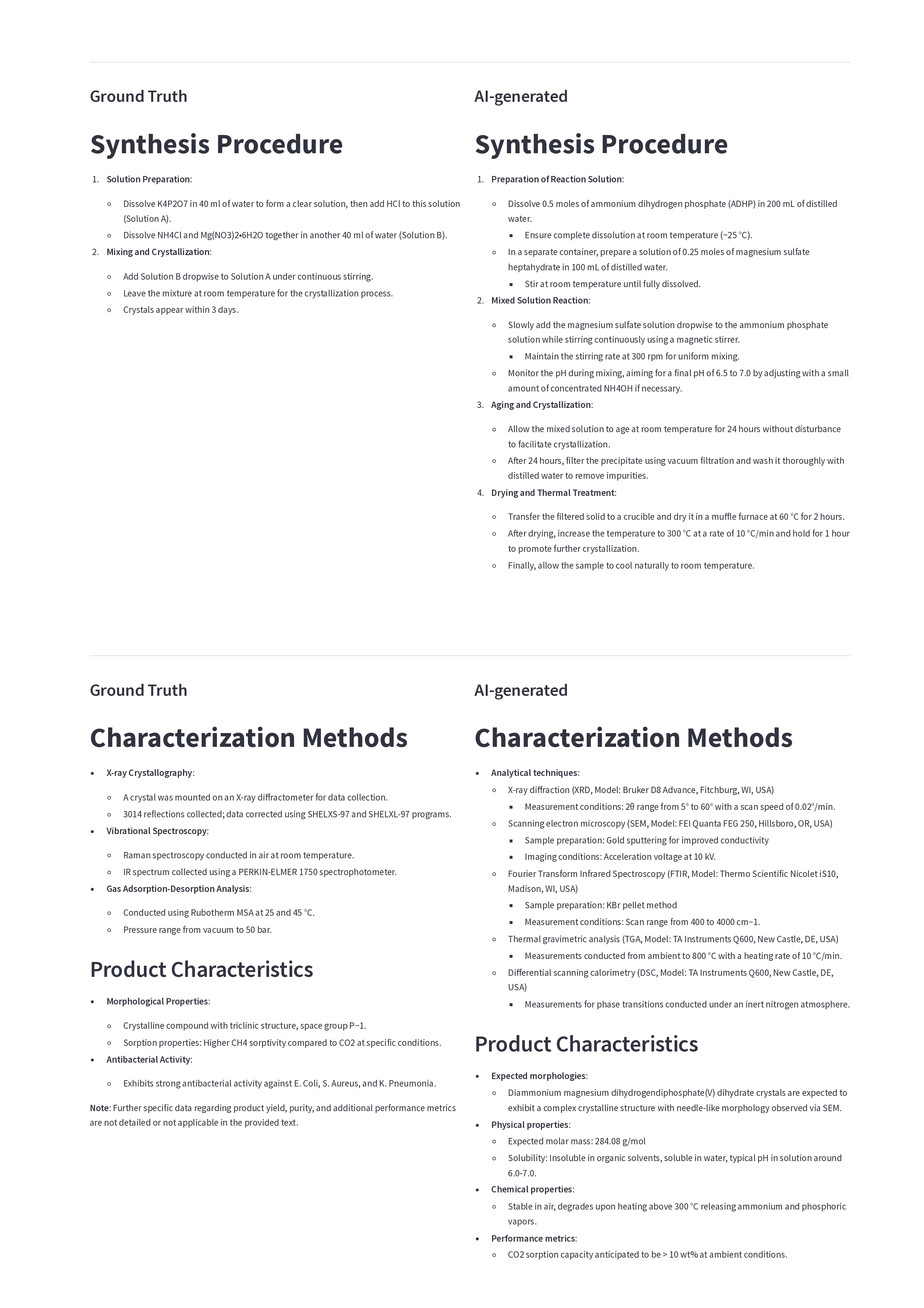}
    \caption{A web UI screenshot for domain experts annotation (2/2).}
    \label{fig:experts-webui-2}
\end{figure*}

Table~\ref{tab:expert_groups} describes the anonymized details about the eight domain experts in materials science. They participated as volunteers and received no evaluation fees.
Figure~\ref{fig:experts-webui-1} and~\ref{fig:experts-webui-2} demonstrate the web UI screenshots for evaluating LLM predictions by domain experts. Domain experts evaluated 20 LLM predictions with seven criteria in Table~\ref{tab:judgment_criteria} and recorded the results in a spreadsheet. We aggregated those eight spreadsheets and calculated the agreement.

\input{table/experts/experts}


\subsection{LLM-Expert Agreement Details}
\label{subsec:llm-expert-agreement-details}

The agreement analysis between expert groups (entire 8-member panel vs. 3-member high-confidence group) and GPT-4o-Nov reveals distinct patterns across evaluation criteria, as shown in Tables \ref{tab:llm-expert-agreement-criteria-full} and \ref{tab:llm-expert-agreement-criteria-high-group}.
These results highlight the critical influence of expert group composition on LLM alignment assessment. Compared to the entire panel, the high-impact subgroup demonstrates enhanced agreement on procedural elements but reduced consensus on characterization tasks, suggesting domain-specific expertise differentially weights evaluation criteria. The stability of non-significant results across both groups for equipment and feasibility judgments implies fundamental challenges in consistently operationalizing these metrics.

\input{table/experts/llm-experts-criteria}

\vspace{2em}
\section{Experimental and Implementation Details}
\label{sec:appendix_experiments}

This section thoroughly describes the LLM prompts and hyperparameter settings to facilitate reproducibility.


\subsection{Hyperparameters}
We use a temperature of zero, top-p of 1.0, and max\_tokens of 4096 for GPT-4o-mini and GPT-4o variants. o3-mini variants use max\_completion\_tokens of 16384, and OpenAI does not allow to set temperature and top-p hyperparameters for o3-mini models.

\subsection{LLM Prompt}
\label{subsec:appendix_llm_judge}
Figure~\ref{fig:prompt_judge} describes the LLM-as-a-Judge prompt. The LLM outputs the JSON-formatted judgment of seven criteria and an overall score for extraction.

\subsection{Other Artifacts}
We utilized LiteLLM~\cite{litellm} and FAISS~\cite{douze2024faiss}, Huggingface Datasets~\cite{lhoest-etal-2021-datasets}.
We confirmed that all models, datasets, and frameworks are allowed for research use.

\vspace{2em}
\section{Additional Results and Analysis}
\label{sec:appendix_results}

Table~\ref{tab:rag_full_result} describes the detailed result of RAG experiments in Figure~\ref{fig:rag_high_impact} for four base LLMs.



\vspace{2em}
\section{Ethical Considerations and Potential Risks}  
\label{sec:appendix_ethics}
Our data collection approach exclusively utilized open-access publications from six major publishers to ensure copyright compliance. 
Additionally, we verified the 12958 articles through keyword-based content filtering and selenium-confirmed articles of CC-BY licensing status, supplemented by a manual sampling of 100 randomly selected articles to validate redistribution rights.
While this strategy mitigates legal risks, two potential limitations warrant consideration: First, the open-access corpus may exhibit selection bias toward well-funded research domains (e.g., energy materials) versus proprietary industrial methods. Second, automated extraction via GPT-4o risks propagating subtle errors from source documents, particularly in stoichiometric ratios and procedural sequencing, despite our expert validation protocol. All dataset derivatives will be distributed under original CC-BY licenses.

\input{appendix_for_submission}

\input{table/experts/rag_full}




\input{appendix_prompts}

%% file: table/data/high_impact_journals.tex
\begin{table*}[!ht]
\centering
\caption{A list of high-impact journals (IF $\ge 10$) that at least ten papers are included in \oursdatashort.}
\label{tab:high_impact_journals}
\adjustbox{max width=\textwidth}{
\begin{tblr}{
  vline{2} = {-}{},
  hline{1,7} = {-}{0.08em},
  hline{2} = {-}{},
}
\textbf{Publisher} & \textbf{Journals}                                                                                                                                                                                                                         \\
ACS                & ACS Applied Materials and Interfaces, ACS Nano, ACS Energy Letters, Journal of the American Chemical Society, ACS Catalysis                                                                                                               \\
RSC                & Journal of Materials Chemistry A, Chemical Society Reviews, Energy  Environmental Science                                                                                                                                                 \\
Nature             & {Nature Communications, Nature Materials, Nature Nanotechnology, Nature Energy, Nature Reviews Materials, Nature Catalysis, \\Nature, Nature Electronics, Nature Methods, Nature Chemistry, Nature Physics, Light: Science  Applications} \\
Wiley              & {Advanced Materials, Advanced Energy Materials, Small, Angewandte Chemie, Advanced Science, ChemSusChem,\\Advanced Functional Materials}                                                                                                  \\
Springer           & Nano-Micro Letters, Journal of Advanced Ceramics, Advanced Composites and Hybrid Materials                                                                                                                                                  
\end{tblr}
}
\end{table*}

%% file: table/experts/experts.tex
\begin{table}
\centering
\caption{Anonymized details for the domain experts in our study. Conf. denotes the average confidence for evaluating LLM predictions in Section~\ref{sec:reliability}. Group C is the highest confidence group.}
\label{tab:expert_groups}
\adjustbox{max width=\linewidth}{
\begin{tabular}{c|l|r} 
\toprule
\textbf{Group} & \textbf{Expertise}                                                                                                                                                                                                  & \textbf{Conf.}  \\ 
\hline
A              & \begin{tabular}[c]{@{}l@{}}One master and two PhD candidates\\specialized in:\\- Thin film transistors\\- 3D nano-semiconductor thin films\\- atomic layer deposition\end{tabular}                            & 1.90                           \\ 
\hdashline
B              & \begin{tabular}[c]{@{}l@{}}Two master's student\\specialized in:\\- Materials modeling\\- DFT \& MD simulations\\- NLP for materials science\end{tabular} & 3.15                           \\ 
\hdashline
C              & \begin{tabular}[c]{@{}l@{}}Three master's student\\specialized in:\\- Transparent electrodes\\- electrospinning\\- 2D materials\end{tabular}                                                        &  \textbf{4.47}                         \\
\bottomrule
\end{tabular}
}
\end{table}

%% file: table/experts/llm-experts-criteria.tex
\begin{table}
\centering
\caption{A agreement between entire expert consensus and GPT-4o-Nov for each criterion. Subscripts denote the $p-$value.}
\label{tab:llm-expert-agreement-criteria-full}
\adjustbox{max width=\linewidth}{
\begin{tblr}{
  column{2} = {r},
  column{3} = {r},
  vline{2} = {-}{},
  hline{1,9} = {-}{0.08em},
  hline{2} = {-}{0.05em},
}
\textbf{Category}                & \textbf{Pearson} & \textbf{Spearman} \\
Material Appropriateness         & 0.59$_{~0.01}$      & 0.59$_{~0.01}$      \\
Equipment Appropriateness        & -0.25$_{~0.29}$    & -0.25$_{~0.28}$     \\
Procedure Completeness           & 0.05$_{~0.83}$     & 0.09$_{~0.71}$      \\
Procedure Similarity             & 0.41$_{~0.07}$     & 0.40$_{~0.08}$       \\
Procedure Feasibility            & -0.04$_{~0.86}$    & -0.04$_{~0.86}$     \\
Characterization Appropriateness & 0.43$_{~0.06}$     & 0.42$_{~0.07}$      \\
Characterization Similarity      & 0.45$_{~0.05}$     & 0.47$_{~0.04}$      
\end{tblr}
}
\end{table}

\begin{table}
\centering
\caption{A agreement between the expert consensus of the high-confidence group and GPT-4o-Nov for each criterion. Subscripts denote the $p-$value.}
\label{tab:llm-expert-agreement-criteria-high-group}
\adjustbox{max width=\linewidth}{
\begin{tblr}{
  column{2} = {r},
  column{3} = {r},
  vline{2} = {-}{},
  hline{1,9} = {-}{0.08em},
  hline{2} = {-}{0.05em},
}
\textbf{Category}                 & \textbf{Pearson} & \textbf{Spearman} \\

Material Appropriateness & 0.44$_{~0.05}$ & 0.41$_{~0.07}$ \\
Equipment Appropriateness & 0.10$_{~0.68}$ & 0.13$_{~0.58}$ \\
Procedure Completeness & 0.23$_{~0.33}$ & 0.20$_{~0.39}$ \\
Procedure Similarity & 0.56$_{~0.01}$ & 0.50$_{~0.02}$ \\
Procedure Feasibility & -0.04$_{~0.86}$ & -0.04$_{~0.86}$ \\
Characterization Appropriateness & 0.43$_{~0.06}$ & 0.42$_{~0.07}$ \\
Characterization Similarity & 0.09$_{~0.72}$ & 0.16$_{~0.50}$ \\
\end{tblr}
}
\end{table}

%% file: appendix_for_submission.tex
\section{AI Assitant}
We use \href{https://copilot.microsoft.com/}{Microsoft Copilot} as a coding assistant and \href{https://grammarly.com/}{Grammarly} and \href{https://www.writefull.com/}{Writefull} as a writing assistant.

%% file: table/experts/rag_full.tex
\begin{table}
\centering
\caption{A full experiment result of RAG prediction in Section~\ref{sec:experiment}}
\label{tab:rag_full_result}
\adjustbox{max width=\linewidth}{
\begin{tblr}{
  row{3} = {r},
  row{4} = {r},
  row{5} = {r},
  row{6} = {r},
  row{8} = {r},
  row{9} = {r},
  row{10} = {r},
  row{11} = {r},
  row{13} = {r},
  row{14} = {r},
  row{15} = {r},
  row{16} = {r},
  row{18} = {r},
  row{19} = {r},
  row{20} = {r},
  row{21} = {r},
  cell{1}{2} = {r},
  cell{1}{3} = {r},
  cell{1}{4} = {r},
  cell{1}{5} = {r},
  cell{1}{6} = {r},
  cell{2}{1} = {r=5}{},
  cell{2}{2} = {r},
  cell{2}{3} = {r},
  cell{2}{4} = {r},
  cell{2}{5} = {r},
  cell{2}{6} = {r},
  cell{7}{1} = {r=5}{},
  cell{7}{2} = {r},
  cell{7}{3} = {r},
  cell{7}{4} = {r},
  cell{7}{5} = {r},
  cell{7}{6} = {r},
  cell{12}{1} = {r=5}{},
  cell{12}{2} = {r},
  cell{12}{3} = {r},
  cell{12}{4} = {r},
  cell{12}{5} = {r},
  cell{12}{6} = {r},
  cell{17}{1} = {r=5}{},
  cell{17}{2} = {r},
  cell{17}{3} = {r},
  cell{17}{4} = {r},
  cell{17}{5} = {r},
  cell{17}{6} = {r},
  vline{3} = {-}{},
  hline{1,22} = {-}{0.08em},
  hline{2,7,12,17} = {-}{},
}
\textbf{Model} & \textbf{K} & \textbf{Mean} & $\sigma$ & \textbf{Min} & \textbf{Max} \\
o3-mini-high   & 0          & 3.759         & 0.407    & 2.86         & 4.71         \\
               & 1          & 3.937         & 0.401    & 2.80         & 4.86         \\
               & 5          & 4.001         & 0.384    & 3.00         & 4.80         \\
               & 10         & 3.939         & 0.359    & 3.00         & 4.80         \\
               & 25         & 3.986         & 0.383    & 3.00         & 4.71         \\
o3-mini-medium & 0          & 3.714         & 0.411    & 2.86         & 4.64         \\
               & 1          & 3.867         & 0.381    & 3.00         & 4.80         \\
               & 5          & 3.934         & 0.349    & 2.86         & 4.80         \\
               & 10         & 3.937         & 0.390    & 2.57         & 4.80         \\
               & 25         & 3.975         & 0.413    & 2.50         & 4.80         \\
o3-mini-low    & 0          & 3.676         & 0.407    & 2.50         & 4.57         \\
               & 1          & 3.848         & 0.411    & 2.90         & 4.80         \\
               & 5          & 3.904         & 0.393    & 2.93         & 4.86         \\
               & 10         & 3.917         & 0.397    & 2.86         & 4.86         \\
               & 25         & 3.961         & 0.388    & 3.00         & 4.80         \\
GPT-4o Nov     & 0          & 3.709         & 0.410    & 2.75         & 4.71         \\
               & 1          & 3.824         & 0.417    & 2.57         & 4.80         \\
               & 5          & 3.949         & 0.391    & 2.93         & 4.80         \\
               & 10         & 3.954         & 0.366    & 3.00         & 4.80         \\
               & 25         & 3.976         & 0.375    & 2.93         & 4.86         
\end{tblr}
}
\end{table}

%% file: appendix_prompts.tex
\begin{figure*}[!h]
    \centering
\begin{lstlisting}
# Solid-State Processing
solid state sintering process, reactive sintering synthesis,
pressure-assisted sintering, spark plasma sintering,
hot pressing synthesis, hot isostatic pressing,
cold isostatic pressing, flash sintering technique,
field-assisted sintering, microwave sintering process,

# Mechanochemical Methods
high energy ball milling, mechanical alloying synthesis,
mechanochemical activation, planetary ball milling,
cryogenic milling process, attrition milling synthesis,
mechanical grinding method, mechanofusion process,
mechano-chemical reaction, solid-state mechanical synthesis,

# Vapor Deposition Techniques 
atomic layer deposition, plasma enhanced CVD,
metal organic CVD, low pressure CVD,
atmospheric pressure CVD, electron beam PVD,
magnetron sputtering deposition, pulsed laser deposition,
thermal evaporation method, molecular beam epitaxy,

# Advanced Thermal Methods
combustion synthesis process, self-propagating synthesis,
plasma spray synthesis, flame spray pyrolysis,
laser ablation synthesis, thermal plasma synthesis,
microwave-assisted synthesis, ultrasonic spray pyrolysis,
radio frequency thermal plasma, arc discharge synthesis,

# Electrochemical Approaches
electrochemical co-deposition, pulse electrodeposition,
electroless deposition method, anodic oxidation synthesis,
cathodic reduction process, electrochemical etching,
electrochemical polymerization, electrophoretic deposition,
galvanic replacement reaction, electrochemical exfoliation,

# Novel Processing Methods
freeze drying synthesis, spray freeze drying,
supercritical fluid process, template-assisted synthesis,
biomimetic processing method, sol-gel electrospinning,
ionothermal synthesis route, microemulsion technique,
sonochemical processing, continuous flow synthesis
\end{lstlisting}
    \caption{60 search keywords to retrieve the literature, including materials synthesis recipes using Semantic Scholar API.}
    \label{fig:prompt_search_keywords}
\end{figure*}

\begin{figure*}[!h]
    \centering
\begin{lstlisting}
Analyze the given scientific text and provide classifications in the following order:

1. Synthesis Recipe Classification:
Determine if the text contains detailed synthesis procedures.
Return only "YES" or "NO".
If "NO", stop here. If "YES", continue with the following classifications.

2. Target Classification:
Classify the synthesized target as one of:
- Material (e.g., nanoparticles, compounds, composites)
- Device (e.g., sensors, batteries, transistors)
- Molecule (e.g., organic compounds, polymers)

3. Material Identification:
Provide:
- Chemical formula (if applicable)
- Material name
- Material class (e.g., metal oxide, polymer, semiconductor)

4. Application Domain:
List the primary applications mentioned in the text:
- Energy (e.g., batteries, solar cells)
- Electronics (e.g., transistors, sensors)
- Healthcare (e.g., drug delivery, imaging)
- Environmental (e.g., catalysis, filtration)
- Others (specify)

5. Synthesis Process Classification:
Classify the given synthesis method into one of these categories. If it combines multiple methods, label it as "Hybrid". If it doesn't fit any category, label it as "Others".

Categories:
1. Solid-State: solid-state reaction, ceramic method, sintering
2. Vapor Deposition: CVD, PVD, sputtering, evaporation
3. Mechanochemical: ball milling, mechanical alloying
4. Hydrothermal: solvothermal, pressurized solution
5. Pyrolysis: thermal decomposition, spray pyrolysis
6. Melt Quenching: rapid solidification, glass formation
7. Electrochemical: electrodeposition, anodization
8. Self-Assembly: molecular assembly, biomineralization
9. Solution-Based: precipitation, sol-gel, wet chemical synthesis
10. Biological: biomimetic, enzyme-mediated, microbial synthesis
11. Hybrid: combination of multiple methods
12. Others: novel or unconventional methods

Format the output as a structured list only if Step 1 is "YES".
For not available, use "N/A".
Do not provide explanations or additional commentary.

Example Output:
For a paper titled "Hydrothermal Synthesis of LiFePO4/C Composites for High-Performance Lithium-Ion Batteries":

1. Synthesis Recipe: YES
2. Target: Material
3. Material Identification:
- Chemical Formula: LiFePO4/C
- Material Name: Carbon-coated lithium iron phosphate
- Material Class: Phosphate composite
4. Application Domain: Energy (lithium-ion batteries)
5. Synthesis Process: Hydrothermal (solvothermal)

Scientific Paper:
{text}
\end{lstlisting}
    \caption{System prompt to categorize the literature converted to markdown format.}
    \label{fig:prompt_paper_categorization}
\end{figure*}

\begin{figure*}[!h]
    \centering
\begin{lstlisting}
You are a materials science expert. Your task is to extract ONLY the explicitly stated synthesis information from the provided research paper. Do not generate, assume, or infer any information not directly presented in the paper.
If the provided paper does not contain any synthesis information, please indicate "NOT A MATERIAL SYNTHESIS PAPER" and do not provide any further details.

## Key Contributions
Summarize the key contributions of the paper:
- Novel materials or compounds: <summary>
- Unique synthesis methods: <summary>
- Specific applications or domains: <summary>

## Materials
Extract and list:
- All precursor materials with:
  * Exact quantities and concentrations
  * Molar ratios or stoichiometric proportions
  * Purity grades and specifications
  * Supplier information if provided
- Solvents, reagents, catalysts, and any other materials such as carrier gases.

## Synthesis Equipment
- All equipment and apparatus with:
  * Model numbers if specified
  * Operating parameters
  * Special configurations or modifications

## Synthesis Procedure
Extract and organize:
- Chronological step-by-step synthesis method
- All processing parameters:
  * Temperature ranges and ramp rates
  * Time durations for each step
  * Pressure conditions
  * pH values if applicable
  * Mixing speeds and durations
- Critical control points and special conditions

## Characterization Methods and Equipment
List all:
- Analytical techniques used
- Specific measurement conditions
- Sample preparation methods
- Equipment models and settings
- Standards or references used

## Product Characteristics
Document:
- Final product properties and specifications (include both numerical values and literal descriptions if provided)
- Yield calculations and actual yields
- Purity levels and impurity content
- Performance metrics with measured values
- Morphological characteristics

IMPORTANT RULES:
1. DO NOT generate or assume any missing information
2. If specific details are not mentioned in the paper, indicate "N/A"
3. Use exact numbers and units as presented in the paper
4. Maintain original measurement units
5. Quote unusual or specific procedures directly when necessary
6. Format all information using proper markdown with headers (##) and bullet points

Remember: Accuracy and authenticity are crucial. Only include information explicitly stated in the paper.

Scientific Paper:
{text}
\end{lstlisting}
    \caption{System prompt to extract the recipe from literature converted to markdown format.}
    \label{fig:prompt_paper_extraction}
\end{figure*}

\begin{figure*}[!h]
    \centering
\begin{lstlisting}[basicstyle=\ttfamily\small]
You are a materials science expert. Your task is to extract ONLY the explicitly stated synthesis information from the provided research paper. Do not generate, assume, or infer any information not directly presented in the paper.
If the provided paper does not contain any synthesis information, please indicate "NOT A MATERIAL SYNTHESIS PAPER" and do not provide any further details.

## Key Contributions
Summarize the key contributions of the paper:
- Novel materials or compounds: <summary>
- Unique synthesis methods: <summary>
- Specific applications or domains: <summary>

## Materials
Extract and list:
- All precursor materials with:
  * Exact quantities and concentrations
  * Molar ratios or stoichiometric proportions
  * Purity grades and specifications
  * Supplier information if provided
- Solvents, reagents, catalysts, and any other materials such as carrier gases.

## Synthesis Equipment
- All equipment and apparatus with:
  * Model numbers if specified
  * Operating parameters
  * Special configurations or modifications

## Synthesis Procedure
Extract and organize:
- Chronological step-by-step synthesis method
- All processing parameters:
  * Temperature ranges and ramp rates
  * Time durations for each step
  * Pressure conditions
  * pH values if applicable
  * Mixing speeds and durations
- Critical control points and special conditions

## Characterization Methods and Equipment
List all:
- Analytical techniques used
- Specific measurement conditions
- Sample preparation methods
- Equipment models and settings
- Standards or references used

## Product Characteristics
Document:
- Final product properties and specifications (include both numerical values and literal descriptions if provided)
- Yield calculations and actual yields
- Purity levels and impurity content
- Performance metrics with measured values
- Morphological characteristics

IMPORTANT RULES:
1. DO NOT generate or assume any missing information
2. If specific details are not mentioned in the paper, indicate "N/A"
3. Use exact numbers and units as presented in the paper
4. Maintain original measurement units
5. Quote unusual or specific procedures directly when necessary
6. Format all information using proper markdown with headers (##) and bullet points

Remember: Accuracy and authenticity are crucial. Only include information explicitly stated in the paper.

Scientific Paper:
{text}
\end{lstlisting}
    \caption{A prompt to predict the recipe with a one-shot example.}
    \label{fig:prompt_prediction_1shot}
\end{figure*}


\begin{figure*}[!h]
    \centering
\begin{lstlisting}
You are an expert in material science and chemical synthesis. Your task is to design a detailed material synthesis recipe, considering the following **key contributions**:
Your output should follow the exact structure and format of the below references. Provide precise details for each section, including materials, equipment, and step-by-step procedures.

Here are recipes from research papers with similar contributions.
-- References --
{references}

-- Prediction Input --
Based on these references, predict the synthesis recipe for the given key contributions:
{contributions}
\end{lstlisting}
    \caption{A prompt to predict the recipe using retrieval-augmented generation}
    \label{fig:prompt_prediction_rag}
\end{figure*}

\begin{figure*}[!h]
    \centering
\begin{lstlisting}
You are an expert materials scientist tasked with evaluating an AI-generated synthesis recipe. You will be provided with:
1. Target material description
2. AI-generated recipe
3. Ground truth recipe from literature

Please evaluate the AI-generated recipe according to the following criteria on a scale of 1-5 (1.0: poor, 5.0: excellent, 0.5 step). Provide detailed justification for each score.

Guidelines for evaluation:

1. Materials
- Appropriateness: Are the selected materials suitable for the target synthesis?

2. Equipment
- Appropriateness: Is the selected equipment suitable?

3. Procedure
- Completeness: Are all necessary steps included with sufficient detail?
- Similarity: How closely does it match the ground truth procedure?
- Feasibility: Can this procedure be realistically executed in a lab?

4. Characterization
- Appropriateness: Are the metrics suitable for validating success?
- Similarity: How well do predicted properties match actual results?

Your Output format must be as following:
1. Step-by-step reasoning first
2. JSON score result
```json
{
    "materials_appropriateness_score": float, 
    "equipment_appropriateness_score": float,
    "procedure_completeness_score": float,
    "procedure_similarity_score": float,
    "procedure_feasibility_score": float,
    "characterization_appropriateness_score": float,
    "characterization_similarity_score": float,
    "overall_score": float
}
```
\end{lstlisting}
    \caption{A prompt to judge the prediction recipe using LLM-as-a-Judge}
    \label{fig:prompt_judge}
\end{figure*}